%% file: acl_latex.tex
\pdfoutput=1

\documentclass[11pt]{article}

\usepackage[]{acl}

\usepackage{hyperref}
\usepackage{times}
\usepackage{latexsym}
\usepackage{tabularx}
\usepackage{lipsum} 
\usepackage{multirow}
\usepackage{array}
\usepackage{booktabs}
\usepackage{makecell}
\usepackage{float}
\usepackage{enumitem}
\usepackage[T1]{fontenc}
\usepackage{graphicx}
\usepackage{rotating}
\usepackage{caption}
\usepackage[table,xcdraw]{xcolor}
\usepackage{colortbl} 
\usepackage{graphicx}
\usepackage{subcaption}
\usepackage[utf8]{inputenc}
\usepackage{tikz}
\newcolumntype{P}[1]{>{\centering\arraybackslash}p{#1}}
\usepackage{xcolor}
\definecolor{forestgreen}{RGB}{34,139,34}

\usepackage{microtype}

\usepackage{inconsolata}

\usepackage{tikz}
\usepackage{tikz-dependency}
\newcommand*\circled[1]{\tikz[baseline=(char.base)]{
            \node[shape=circle,draw,inner sep=.6pt] (char) {#1};}}

\title{"I Never Said That": A dataset, taxonomy and baselines on response clarity classification.}

\author{Konstantinos Thomas\textsuperscript{1}
  Giorgos Filandrianos\textsuperscript{1}
  Maria Lymperaiou\textsuperscript{1} \\
  \textbf{Chrysoula Zerva\textsuperscript{2,3,4}}\and
  \textbf{Giorgos Stamou\textsuperscript{1}} \\ 
  \textsuperscript{1}National Technical University of Athens \\
  \textsuperscript{2}Instituto de Telecomunicações \\
  \textsuperscript{3}Instituto Superior Técnico, Universidade de Lisboa\\
  \textsuperscript{4}ELLIS Unit Lisbon \\
  \texttt{\{kthomas, geofila, marialymp\}@islab.ntua.gr} \\
  \texttt{chrysoula.zerva@tecnico.ulisboa.pt, gstam@cs.ntua.gr }
  }
\begin{document}
\maketitle 
\begin{abstract}
Equivocation and ambiguity in public speech are well-studied discourse phenomena, especially in political science and analysis of political interviews. Inspired by the well-grounded theory on equivocation, we aim to resolve the closely related problem of response clarity in questions extracted from political interviews, leveraging the capabilities of Large Language Models (LLMs) and human expertise. To this end, we introduce a \textit{novel taxonomy} that frames the task of detecting and classifying response clarity and a corresponding \emph{clarity classification dataset} which consists of question-answer (QA) pairs drawn from political interviews and annotated accordingly. Our proposed two-level taxonomy addresses the clarity of a response in terms of the information provided for a given question (high-level) and also provides a fine-grained taxonomy of evasion techniques that relate to unclear, ambiguous responses (lower-level). 
We combine ChatGPT and human annotators to collect, validate and annotate discrete QA pairs from political interviews, to be used for our newly introduced response clarity task.
We provide a detailed analysis and conduct several experiments with different model architectures, sizes and adaptation methods to gain insights and establish new baselines over the proposed dataset and task.~\footnote{Code and data can be found here: \url{https://github.com/konstantinosftw/Question-Evasion}.}

\end{abstract}

\section{Introduction}

\begin{figure}[h!]
    \centering
    \includegraphics[width=\columnwidth]{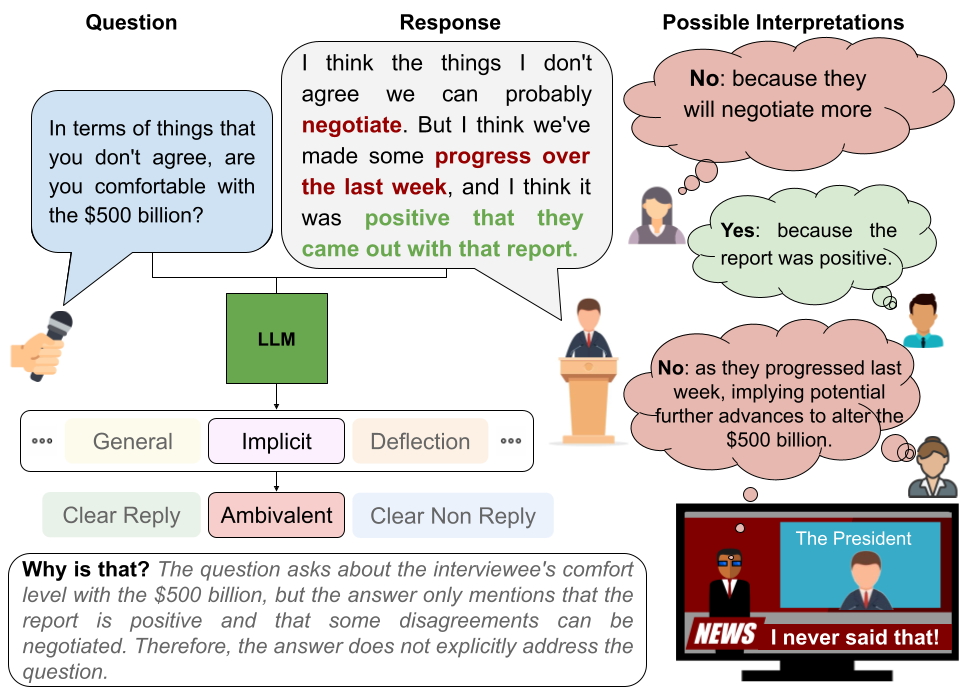}
    \caption{
    An example from an interview from our dataset with classification along with an analysis from instruction-tuned Llama-70b.}
    \label{fig:graph}
\end{figure}

\begin{figure}[h!]
    \centering
    \vskip -0.05in
    \includegraphics[width=0.95\columnwidth]{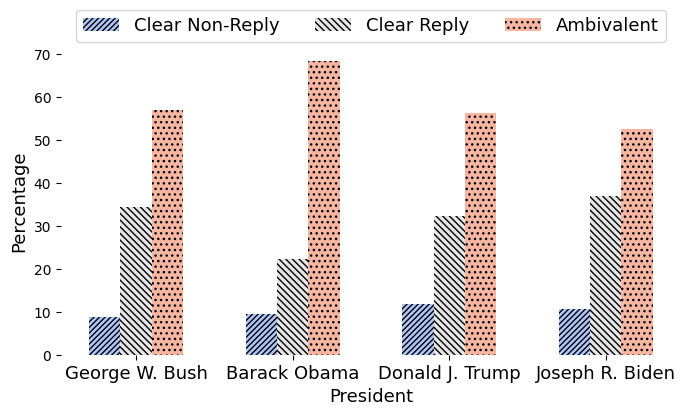}
    \caption{Statistics on answer clarity in political interviews of the latest 4 US presidents.}
    \label{fig:barplot_pres}
\end{figure}

In the era of mass information dissemination, question evasion and response ambiguity are widespread phenomena in political interviews and debates, rendering their detection an important aspect of political discourse studies. \citet{Bull2003PoliticalCommunications} presents a meta-analysis of five studies on political interview Q\&As, concluding that politicians gave clear responses to only 39-46\% of questions during televised interviews, while non-politicians had a significantly higher 70-89\% reply rate. In Figure \ref{fig:barplot_pres} we present statistics derived from our human annotations regarding response clarity among US presidents, revealing that politicians often avoid providing clear responses to journalists' questions. 

This phenomenon is known as \textit{equivocation} or \textit{evasion} in academic literature and describes a non-straightforward type of communication, which is characterised by lack of clarity and includes speech acts such as contradictions, inconsistencies, subject switches, incomplete sentences, misunderstandings, obscure mannerisms of speech \cite{Watzlawick1964PragmaticsOH, Bavelas1988PoliticalEA, Rasiah2010AFF}, rendering political speech susceptible to multiple interpretations from the perspective of the public. Figure \ref{fig:graph} presents an example of an interview featuring various interpretations, generated labels, and corresponding explanations using our proposed dataset.

While the topic has been studied extensively in the field of linguistics, politics and communication, with several typologies proposed for classifying question responses \cite{Harris1991, Bull1993HowNT, Rasiah2010AFF}, there has been little attempt to analyse whether such typologies are applicable to larger scale data and consistent with varying human perspectives and biases. In other words, the possibility of automatically classifying response clarity has not been explored in NLP, potentially due to the complexity of the task itself, as well as the underlying need to encode and reason over 
long context. However, recent advancements in language modelling boosted model performance for long-context inputs \cite{dai2019transformer,wei2022emergent,wei2023larger}, paving the way for framing the task of \textbf{automatically measuring response clarity}. 


Related to this endeavour, there is related work focusing on the responder's intent interpretation \cite{ferracane-etal-2021-answer}, or the answerability of questions for question-answering (QA) tasks \cite{min-etal-2020-ambigqa, bingningwang2020reco, quail, sun-etal-2022-conditionalqa, wang2022archivalqa}. However, in both research directions, the focus deviates from directly assessing the clarity of the response, being obfuscated by perceptions of intent or question clarity. We address this by proposing the task of \textbf{response clarity evaluation}, focusing exclusively on assessing the effect of the response, building on relevant discourse typologies.

We carry out a detailed analysis of proposed typologies, considering their overlap and consistency, the distribution of proposed classes in our collected data, and the feasibility of using them in an automated task, resulting in our proposed \textit{two-level response clarity detection taxonomy}. Specifically, the first level of the taxonomy accounts for a three-way evaluation of response \textit{clarity} in terms of the number of interpretations the intended response holds. The second and more fine-grained level covers eleven common \textit{evasion} phenomena in political literature, which explain in more detail the categorization of responses in the three-scale clarity classes. 
We use this taxonomy to annotate a dataset of political QA pairs and perform an analysis of the perspective variability among human annotators. We then evaluate different LLMs, exploring various training and inference frameworks, showing that simple prompting and instruction-tuning techniques using our dataset are highly capable of providing meaningful performance. Moreover, we find that using the labels of the second level (evasion labels) in a two-step classification strategy helps boost performance for clarity classification.

We argue that being able to detect answer ambiguity automatically will facilitate political speech discourse analysis, allowing for comparisons at scale. Additionally, the proposed task can shed light on LLM capabilities of reasoning over long contexts and prove useful for other downstream tasks in NLP such as question answering (see also §\ref{sec:rw_social}).
To sum up, our contributions are threefold:
\begin{itemize}\setlength\itemsep{0em}
\item We propose a new task, \textit{response clarity evaluation}, which aims to detect the alignment and clarity of a given response with respect to its respective question and provide an empirically and theoretically established taxonomy for it.
\item We introduce a human-labelled dataset on the aforementioned task, comprising 3,445 QA pairs from political interviews. 
\item We experiment with several language models and methods to gain insights establish performance baselines for the proposed task.
\end{itemize}

\section{Related work}
\subsection{Equivocation in Social Sciences}
\label{sec:rw_social}

Political equivocation, aptly generalised by \citet{Dillon1990ThePO} as ``the routine strategy for responding to a question without answering it'', provides a range of frameworks to analyse evasive responses 
\cite{Wilson1990PoliticallyST, bull2009, Bull2019CantAW}. 
\citet{Harris1991} makes a distinction between direct and indirect answers while others focus on how complete the information conveyed by the response is \cite{Bull1994OnIQ, Bull2003PoliticalCommunications}.
\citet{Wilson1990PoliticallyST, Harris1991, Bull2003PoliticalCommunications}  provide criteria for the identification of three main categories \cite{Bull1993HowNT}: \circled{1} \textit{Replies} correspond to cases where the requested information is given in full.  \circled{2} \textit{Non-Replies}, where none of the information requested is given in a clear manner \cite{Rasiah2010AFF}; non-Replies are broken down into twelve further \textit{evasion} sub-categories (Table \ref{table:equivocation_typology}). Lastly, \circled{3} \textit{Intermediate replies} are those utterances that fall somewhere between replies and non-replies, i.e. responding completely but to one part of a multi-part question while ignoring the rest; responding partially to a single-part question; answering a question in a suggestive manner without giving a straightforward answer. 

\citet{Bull2003PoliticalCommunications} breaks the 12 evasion techniques of Table \ref{table:equivocation_typology} further into 28 more fine-grained micro-categories; for example ``\textit{Makes political point}'' includes the micro-categories ``\textit{External attacks on the opposition or other rival groups}'',  ``\textit{Talks up one’s own side}'', ``\textit{Presents policy}''.
\citet{Rasiah2010AFF} separates the Replies into Direct and Indirect, keeps the Intermediate Replies category as is, while also breaking down the Non-reply category (which he labels ``Evasions'') into four degrees of evasiveness, whether the evasion was overt or covert and what types of ‘agenda shifts’ occurred.
\input{tables/rw_evasion_typology}

Tailoring these typologies into a response clarity taxonomy suitable for an NLP dataset, it is imperative to modify them considering the following:
\begin{itemize}\setlength\itemsep{0em}
    \item Our focus is slightly different: we target a taxonomy that classifies the clarity of responses (hence an indirect response falls under a different category than a direct one). 
    \item We seek a good per class representation in our dataset to allow computational modelling using LLMs. It is thus necessary to condense classes to avoid overly sparse categorisation while retaining the essential per class characteristics (i.e., we provide meaningful labels).
    \item Labelling of the responses is conducted by non-expert human annotators so that our annotations also account for the perception and reasoning of the general audience of political interviews rather than a minority of experts. The difficulty of the classification, and thus the resulting error rate, increases as we increase the set of labels they choose from.
    \item Most interviewers pose multi-barrelled questions. We break those multi-part questions into singular QA pairs and label each one separately, to retain this fine-grained information.
\end{itemize}

Section \ref{sec:reply_custom_typology} discusses the taxonomy we adopted, aiming to optimise for the annotation task.

\subsection{Equivocation in NLP} 

While \textit{equivocation} has not been adequately studied in NLP, there are related areas, such as question answerability,  
political discourse analysis and deceptive intent detection.
\begin{figure*}[ht!]
\vskip -0.06in
\centering
    \includegraphics[width=0.8\textwidth]{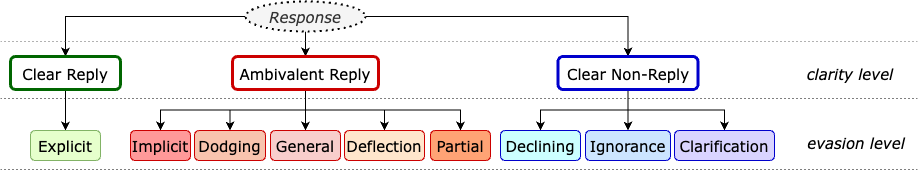}
\vskip -0.01in
\caption{Our proposed taxonomy of response clarity classification.}
\label{fig:typology}
\end{figure*}
\subsubsection{Answerability in question answering}
There have been several tasks proposed related to QA both in open-ended and closed set answer setups. The issue of the \textit{answerability} of a given question an
in QA was highlighted in SQuAD 2.0 \cite{rajpurkar-etal-2018-know}, which introduced adversarially crafted unanswerable questions with respect to a given text span. \citet{lee2020squad2} expanded the SQuAD 2.0 dataset, also incorporating the rationale for unanswerable questions. Extending to out-of-domain questions to address practical use cases, \citet{Sulem2021DoWK} introduce competitive and non-competitive unanswerable questions. Relevant endeavours question the answerability of information-seeking queries built independently of the passage containing possible answers to those queries \cite{Asai2020ChallengesII}. Scalability issues are addressed via synthetic extensions of existing datasets containing both answerable and unanswerable questions \cite{Nikolenko2020WhenID}. To the same end, other works develop data augmentation techniques to produce unanswerable queries based on answerable SQuAD 2.0 queries \cite{Zhu2019LearningTA,Du2022AFF}. Other datasets targeting answerability issues are ReCO \cite{bingningwang2020reco}, which provides ``yes'', ``maybe'' and ``no'' labels for questions paired with passages in Chinese, as well as QuAIL \cite{quail}, which introduces questions of varying certainty according to the accompanying passage. 


While our task shares a connection with question answerability, our focus is on annotating \textit{response clarity} in relation to a given question. This distinction shifts the goal from evaluating \textit{question} clarity leading to a unique task and reasoning process.

\subsubsection{Discourse analysis of political speech}
Beyond evasion, discourse phenomena in political speech (including interview responses) have been analysed in prior NLP works. \citet{majumder-etal-2020-interview} construct a large-scale dataset of political dialogues to study discourse patterns, upon which they train a model that uses external knowledge. Among the analysed discourse patterns they consider modes of persuasion, entertainment, and information elicitation (the latter being the closest to our target). Understanding political agendas requires contextualization, depending on which politician expresses a certain claim: \citet{pujari-goldwasser-2021-understanding} propose the combined use of transformer-based modules to obtain better representations of political agendas based on politician tweets. Finally, non-verbal aspects of political discourse, such as the usage of gestures have been proven to be associated with individuals rather than political parties, while contributing to emphasising certain parts of speech \cite{trotta-tonelli-2021-gestures}.


%
Another relevant dimension that has been explored in the context of automated discourse analysis is detecting the intent of the responder. \cite{girlea_2017} trained Relational Dynamic Bayesian Networks on psycholinguistic features of non-political dialogues to identify linguistic cues associated with deception.
In a work lying closer to ours, \cite{ferracane-etal-2021-answer} crowdsourced annotators to label political interview answers, firstly as "answer", "shift" or "didn't answer" and ultimately whether that act had honest or deceptive intent. They thus aim to collect diverse, subjective opinions on the (dis)honesty of responders providing a valuable two-way view on the topic that involves both the responder and the audience (annotator). We instead opt for avoiding assumptions on speaker intent, and focusing only on discourse techniques the speaker used, since they are better defined in related literature, and allow us to directly evaluate the clarity of a response. For example, an on-topic response that is slightly open to interpretation would be labelled as "Implicit reply" under the "Ambivalent reply" category by our typology. While for \cite{ferracane-etal-2021-answer}, this would fall under the parent category of "Answer", and either "direct" or "overanswer", depending on whether the annotator felt that the speaker was \textit{purposefully} ambiguous or not. This decision on the annotation focus allows us also to annotate a more extensive dataset ($\approx3.4K$ pairs) due to its less subjective nature, which considers the level of clarity and completeness of responses.
 

\section{Proposed Taxonomy
}  \label{sec:reply_custom_typology}



The typologies discussed in §\ref{sec:rw_social} are comprehensive and well-researched, but often exhibit compatibility issues  \cite{Bull1994OnIQ, Bull2019CantAW, Rasiah2010AFF} as distinctions between categories vary among experts and sub-domains. For instance, a somewhat vague reply may be deemed as evasive by some while indirect yet coherent by others, especially since ambivalent responses are particularly prone to confirmation bias \cite{Nickerson1998ConfirmationBA}. To enhance objectivity, we focus on the Clarity/Ambiguity dimension, rather than a Reply/Non-reply distinction. This approach shifts annotators' attention from the bias-prone task of trying to decipher if an answer is ``valid'' or ``invalid'', to whether a response can be interpreted unambiguously or accepts a wider range of interpretations.

Extensive typologies such as \citet{bull2009} include over $30$ types of replies, resulting in a sparse dataset with few examples per category that further complicates the annotation task. 
We thus aimed to consolidate these typologies into fewer essential categories, while maintaining crucial distinctions.

Another necessary adjustment involved breaking down multi-part questions into their constituent questions, which led to the elimination of the category of ``intermediate replies''. As discussed in §\ref{sec:rw_social}, most interviewers pose multi-barrelled questions and vagueness in a single answer towards a multi-part question results in classifying the entire response as an intermediate reply. To avoid skewing the dataset towards intermediate replies, we broke multi-barrelled questions into separate questions and asked the annotators to label each sub-question and answer separately.


Taking all of the above into consideration, we arrived at a two-level hierarchical taxonomy. The higher level includes 3 main response categories, namely \circled{1} \textit{Clear reply}, containing replies that admit only one interpretation; \circled{2} \textit{Clear non-reply}, containing responses where the answerer openly refuses to share information, and \circled{3} \textit{Ambivalent reply}, where a response is given in the form of a valid answer but allows for multiple interpretations. At the second level these 3 categories further split into 9 sub-categories illustrated in Figure~\ref{fig:typology}. As a brief exemplification, ``\textit{Q: Have you seen my chocolates?
A: The children were in your room this morning.}'' would be considered an \textit{Implicit} reply (under the \textit{Ambivalent} category) since there is a rather clear implication on the culprit. Yet, the answer does not commit to explicitly stating that ``the kids ate it'' - which would have made for an Explicit reply - but rather prompts for a reasoning step to reach the final assumption. Instead, ``\textit{A. I don't know}'', for the same question, would be labelled as a \textit{Clear non-reply} and specifically \textit{Claims ignorance}, since the respondent explicitly refuses to provide information; also, ``\textit{A. You should not keep your chocolates all around the house}'' would be considered a \textit{Deflection}, i.e. an \textit{Ambivalent} answer, as it provides none of the requested information, yet it leverages the subject to pivot on a different point. For further analysis and examples see Table~\ref{tab:typology-examples} in App.~\ref{app:examples_taxonomy}.

\section{Dataset creation}
As a first step, we collect presidential interviews of US Presidents, provided by the official Whitehouse website~\footnote{
Interviews from
\href{https://www.whitehouse.gov/}{https://www.whitehouse.gov/}.}. This resulted in 287 unique interviews spanning from 2006 until 2023 which we further analyse in App.~\ref{sec:interviews}. We extracted a total of 3,445 questions and responses from these interviews, as described in the following sections.

We leverage ChatGPT to decompose the original interviews into QA pairs, aiming to separate multi-barrelled questions into separate sub-questions and their respective response sub-parts. We use the automatically generated list of (sub-)questions to generate annotation instances, and then, upon validating the decomposition, annotators label the response to each sub-question separately. Thus, for a given interview question, we may have several QA instances in the final dataset corresponding to distinct sub-questions, and the classification of the respective sub-responses. We henceforth refer to the generated sub-questions and sub-responses as \textit{singular QA pairs}, ``\textit{sQAs}'' for short.



\paragraph{Human annotation process}

Upon the aforementioned preprocessing of the interview questions, we specify the annotation task where the annotators are provided both with the original QAs as well as the decomposed {sQAs}, and asked to label the response for each sub-question separately.  We opted for providing the sQAs alongside the full text to reduce the effort of manually extracting distinct sQAs from the original interviews, which would significantly increase the annotation time per sample. 
We further introduce \textit{counterfactual sQAs} to measure the annotators' potentially exclusive reliance on sQAs, as explained in App. \ref{sec:countersummaries}. We were thus able to verify that all annotators followed our instructions and the introduction of sQAs aids instead of hindering the annotation process. The prompt provided to ChatGPT to create the original sQAs and counterfactual sQAs is shown in App.~\ref{sec:prompting}. 

We employ 3 human annotators alongside an expert with a background in political science and political discourse analysis who acts as a validator of the outcome annotations.
As a first ``training'' stage, we provide the annotators with a tutorial that includes annotated examples from each category of the taxonomy to allow them to familiarise themselves with the concepts introduced. Then, the annotators are prompted to perform a series of annotation tasks in the following order:
they have to \circled{1} evaluate the sQAs produced by ChatGPT as valid or not, and then \circled{2} label each of the individual questions and answers, using the proposed taxonomy or indicate an erroneous question in sQAs. Finally, they should \circled{3} add any missing questions, as well as the corresponding label. 
On average, each annotator evaluated 1150 samples. 
More information is provided in App.~\ref{sec:annotators}.

\begin{figure}[h!]
    \centering
    \includegraphics[width=0.9\columnwidth]{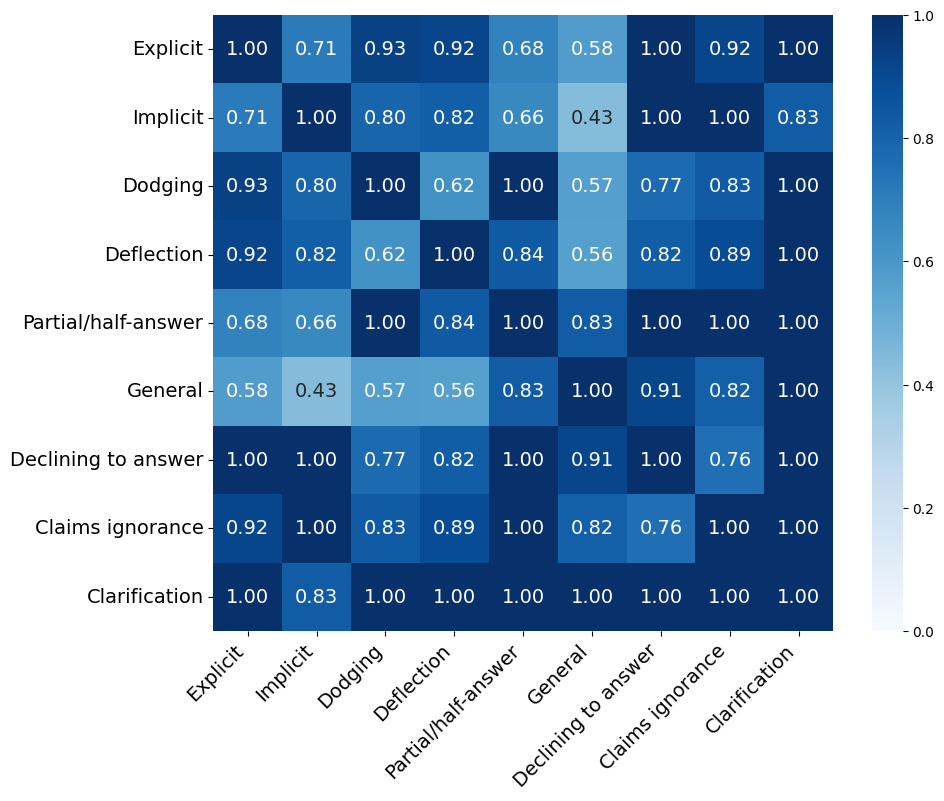}
    \vskip -0.04in
    \caption{Annotators' agreement using Fleiss $\kappa$ for labels assigned to the `evasion' classification  level.}
    \label{fig:heatmap}
\end{figure}

\paragraph{Validation set \& inter-annotator agreement} 
\label{sec:iaa}
As the proposed task is 
challenging and annotator perspectives may influence their decisions, we use a subset of the data ($317$ common QA pairs) as \textit{validation} for which we collect overlapping annotations from all $3$ non-expert annotators. 
We calculate the inter-annotator agreement between the non-experts, for both the fine-grained `evasion' taxonomy categories (Figure \ref{fig:typology}, lower level classes) and the higher-level `clarity' categories. 
We thus aim to both confirm the validity of our annotations and explore which labels draw more disagreements, potentially being more dependent on diverging perspectives and biases of annotators or being inherently harder to distinguish. Table \ref{tab:fleiss-binary} shows the annotators' agreement via Fleiss Kappa $\kappa$ scores \cite{fleissk}  when given samples from two different `clarity' classes (row, column). Similarly,  Figure~\ref{fig:heatmap} concerns the `evasion' level classification.

\begin{table}[h!]
\begin{tabular}{p{1.75cm} P{1.34cm} P{1.75cm} P{1cm}}
\hline
         & \small{Clear R.} & \small{ Clear Non-R.} & \small{Ambiv.}  \\ \hline
\small{Clear R.}   &   \small 1      &  \small   0.97       &     \small   0.65            \\ 
\small{ Clear Non-R.}   &   \small 0.97       &       1      &    \small 0.71            \\   
\small{Ambiv.}   &   \small 0.65       &       \small 0.71       &       \small 1      \\   \hline
\end{tabular}
\caption{Fleiss $\kappa$ (higher values are better) between annotators for the `clarity' classification level.
}
\label{tab:fleiss-binary}
\end{table}

For the `clarity' category, the Fleiss Kappa $\kappa$ indicates moderate to high agreement among non-expert annotators at 0.644, compared to 0.48 for the more challenging `evasion' classification, signifying moderate agreement.
There is near perfect agreement between annotators regarding Clear Reply and Clear Non-Reply ($\kappa$=0.97), while, rather intuitively, confusions occur when distinguishing between \textit{Ambivalent} category and any of the rest. Figure \ref{fig:heatmap} sheds more light on the confused labels
: it seems that annotators diverge more when discriminating between 
\textit{General} (\textit{Ambivalent}) vs \textit{Explicit} (\textit{Clear Reply}) ($\kappa$=0.58) and \textit{Partial} (\textit{Ambivalent}) vs \textit{Explicit} (\textit{Clear Reply}) ($\kappa$=0.68), or
`Declining' (Clear Non-reply) vs`Dodging' (Ambivalent) ($\kappa$=0.77).
On the contrary, there is a clear distinction between `Claim ignorance', 'Decline to answer' `Clarification' categories and 'Explicit' replies ($\kappa\geq$0.92). Moreover, there is also high disagreement within Ambivalent labels, such as 'General' vs 'Implicit', 'General' vs 'Deflection', and
`General' vs `Dodging' categories.

\paragraph{Handling disagreements} 
As we intend to use the described \textit{validation} set in the evaluation stage (i.e. as our test set), we opt for resolving the disagreements and obtaining a single gold label for all these 317 \textit{validation} samples. 
When a disagreement between non-expert annotators occurs, a majority voting scheme is employed to decide the gold label. If there is no majority label, the expert annotator resolves the conflict by assigning the final gold label to the respective samples.

Notably, deviating annotations are not necessarily invalid and can represent a variability of perspectives that could be useful to model instead of resolve. Recent work has highlighted the importance of access to multiple perspectives for complex NLP tasks, encouraged by the emergence of datasets that maintain several annotations per instance to motivate training models under uncertainty or annotation variation  \cite{baan-etal-2022-stop,baan2023uncertainty,plank2022problem,giulianelli-etal-2023-comes}. Hence, and while capturing diverting perspectives is out of scope for this work, we release the full annotations alongside the single-label dataset, to allow for future research into models that can address multi-label scenarios. 

\paragraph{Exploratory data analysis} 
revealed shifts in evasion patterns, such as an increased reply rate at the end of the presidential service for some presidents (e.g. D. Trump), while the opposite behaviour is derived for others (e.g. G. Bush). Additionally, evasion correlates with the presence of multi-part questions 
Interestingly,
while in joint interviews, presidents tend to alter their reply strategy compared to when being interviewed on their own.
We provide more details in App. \ref{sec:interviews}. 

\section{Experiments}
\subsection{Experimental setup}
We test various models on our disagreement-resolved \textit{validation} set to showcase the impact of different modelling choices and establish baselines. Details regarding experiments in App. \ref{sec:exp-details}.

\paragraph{Modeling variants} We compare (i) encoder models: DeBERTa \cite{he2021deberta}, RoBERTa \cite{liu2019roberta}, and XLNet \cite{yang2019xlnet}; (ii)LLMs: Llama2 \cite{llama}, Falcon \cite{falcon}; and (iii) ChatGPT (gpt3.5\textunderscore
turbo)~\footnote{Specifically, we used version gpt-3.5-turbo-0613.}. Additionally, we compare varying adaptation strategies, namely inference via zero (ZS) or few-shot (FS) and chain-of-thought (CoT) prompting variants (prompts provided in App.~\ref{sec:prompting}), as well as instruction-tuning on the target labels using LoRA tuning (more details in App. \ref{app:lora}).

Our CoT approach employs a breakdown of instructions, as well as the ``Let's think step by step'' phrase \cite{kojima2022large}, asking the model to first reason about QAs and then classify based on the taxonomy. We compare two CoT flavors: \circled{1}~\textit{standalone CoT} classifies only one sQA at a time, and \circled{2} \textit{multiple CoT} attempts to classify all sQAs pertaining to a multi-barrelled question in one go.
For the instruction-tuning part, we rely on LoRA fine-tuning~\cite{hu2021lora}. The details of the experiments are provided in App. \ref{sec:exp-details}, while the instruction format is outlined in App. \ref{app:lora}.


\paragraph{Classification variants}
We explore two different classification variants to evaluate responses:
\circled{1}
\textbf{Direct clarity classification}: we tune and prompt models to directly predict one of the 3 labels of the clarity level: Clear reply, Ambivalent Reply and Clear non-reply.
\circled{2}
\textbf{Evasion-based clarity classification}: we infer the clarity labels in two steps. First, we tune and prompt the models to predict the 9 evasion sub-categories (leaves of the taxonomy tree) and then we infer the 3 labels by traversing the taxonomy hierarchy upwards.

\subsection{Results and Discussion}
Classification results for different training and inference strategies are provided in  Table \ref{tab:llm-results-zs}. More detailed analysis can be found in App.\ref{sec:encoder-classification} \footnote{Note that results for XxBERTa models are overestimated due to constraint input token size.}.

\input{tables/zeroshot}

For the \textbf{ZS} setup, we exclusively present results for the larger models due to the very low performance of the smaller ones (Llama 7B/13B and Falcon 7B), which frequently hallucinated and rarely predicted labels within the taxonomy. ChatGPT significantly outperforms the other two models across metrics for both classification variants, and it is positively influenced by the two-step evasion-based strategy. While Falcon also benefits from generating fine-grained labels, Llama exhibits the opposite behaviour, performing worse on the 9-way classification task and thus moving up in the hierarchy leading to increased misclassifications. Instead, Llama has a better representation of the high-level labels, performing better on the direct clarity classification.
For \textbf{FS}, due to the lengthy sQAs of our dataset's interviews, we employ shorter representative examples (Table \ref{tab:typology-examples}).
FS showcased advanced results compared to ZS, with smaller models experiencing a significant reduction in hallucinations. Further analysis is provided 
 in App. \ref{sec:few-shot}.

\textbf{CoT} experiments exhibit a different behaviour for each classification variant. Specifically, CoT improves the performance for the evasion-based strategy only, hinting that the ``step-by-step'' reasoning process is more meaningful when addressing a task with higher dimensionality/complexity of targeted labels. Interestingly, asking to address all sQAs in one go (multi-CoT) harms performance instead of improving, potentially because of the impact on the amount of context that needs to be taken into account for generation.

In general, LLMs mostly struggled with distinguishing between \textit{Clear} vs \textit{Ambivalent} replies, as well as \textit{Partial} vs \textit{General} ones. This resembles challenges (Figure \ref{fig:heatmap}) faced by human annotators but interestingly holds even for ZS and CoT models which were not trained on human annotations, suggesting a generalised difficulty in discerning these classes. Further insights are shown in App. \ref{sec:evasion-classification}.




Turning to tuned models, we observe a difference in behaviour: for direct clarity, smaller LLM models seem to struggle and are even outperformed by encoder models such as XLNet or BERT variants, with only the 70b Llama outperforming them. Instead, evasion-driven classification consistently improves the performance of Llama variants. Additionally, Llama models outperform Falcon even with fewer parameters (e.g. the 13B Llama model outperforms the 40B Falcon across metrics). This aligns with other works where LLama-13b surpasses Falcon-40b in reading comprehension \cite{llama}, while all LLama variants exhibit better prior knowledge \cite{sun2023head}, a crucial factor for our task as discussed below. We expand our experiments to assess the generalisation capabilities of the stronger Llama model (70B) using the dataset of \cite{ferracane-etal-2021-answer}, which is annotated with a different strategy, and provide an analysis as detailed in App. \ref{app:comparison_intent}.

Overall, for both prompting and tuning strategies, the evasion-based clarity classification variant leads to better performance compared to the direct clarity one, indicating that the fine-grained subcategories of the taxonomy assisted in guiding the LLMs towards selecting the correct high-level clarity category more frequently. 
In other words,  while the 9-way classification is more challenging (see also App. \ref{sec:evasion-classification}), disambiguation between the finer-grained labels helps the models improve their accuracy on the higher-level ones. Further analysis of performance per class is provided in App. \ref{app:per_per_class}. 

\paragraph{Answer grounding}
We aim to separately assess whether models are influenced by the difficulty of identifying the relevant response snippets in the text, i.e. grounding the answer, a task that can be particularly challenging when a single reply addresses multiple questions. 
 As a proxy to test this, we consider single- vs multi-part question subsets (35\% vs 65\% of the original test-set), assuming that answer grounding is harder for the latter, and we compare models and annotator performance. While Fleiss $\kappa$ showed minimal disparity between humans across all models, metrics were notably higher for single-part questions, regardless of the method (ZS/FS, CoT, fine-tuning) or the classification variant (evasion-based or direct clarity). Performance improvements reached $0.16$ for F-score, indicating the impact of QA complexity on model performance. 
More detailed results in App \ref{sec:ans_ground}. 

\paragraph{Model knowledge}
We explore whether performance in the proposed task is influenced by models' ``prior knowledge'' of given entities. For instance, Q: ``Did the Federal Reserve make the right move?'', A: ``I think Bernanke is doing a great job'' would be correctly classified as \textit{Dodging} by models unaware that Bernanke is the chairman of the Federal Reserve. To explore the prior knowledge hypothesis, we focus on person names and divide the test-set into two parts: one containing person names in either the question or the answer, and one excluding any named person mentions (60\% vs 40\% of the original test-set). All models performed better on the latter, ``no-person'' subset, but smaller models exhibited a much sharper improvement of up to $0.20$ in F-score (Llama-7b) compared to larger and presumably more ``knowledgeable'' ones, thus corroborating the findings of \citet{sun2023head}. We provide more details in App \ref{sec:prior_knowledge}.

\section{Conclusion}
We introduce a novel task on response clarity classification focusing on political interviews. Driven by studies of evasion techniques in political sciences, we propose a two-level hierarchical taxonomy for clarity classification that considers different evasion strategies at the lower (leaf) level. We also introduce a new dataset where question-answer pairs are manually annotated with the proposed taxonomy labels. We experiment with a range of different model architectures, sizes and adaptation strategies on our dataset, establishing several baselines. We empirically show that fine-grained labels facilitate classification in response clarity, while encoded model knowledge is strongly associated with classification performance. We aspire for this work to motivate future research in the topic,  both from the NLP and political sciences communities.


\section*{Limitations}
Due to the usage of Large Language Models (ChatGPT) in our pipeline, our annotation process is susceptible to hallucinations, which could affect the quality of the sQA extraction and therefore the assignment of correct labels. However, we attempt to mitigate this risk by asserting that our human annotators are attentive and not influenced by injecting counterfactual sQAs. Additionally, we manually inspected the quality of both the ChatGPT-generated sQAs and the human annotations throughout the annotation campaign to ensure high-quality annotations. 
Further, despite being crucial for the quality of the derived dataset, the need for human annotators significantly limits the number of samples that can be annotated, especially when considering the complexity of the proposed task. Overall, our dataset and respective analysis are limited to the English language and further work would be needed to generalise the findings to other languages, especially low-resource ones. Finally, the inherently missing vocal features present in speech, as well as face movements and hand gestures limit the discourse analysis to purely textual cues, potentially missing some evasion-related characteristics.

\section*{Potential risks}
Potential risks associated with this work relate to the possibility of misclassification of a part of political speech due to the usage of neural models (LLMs) as classifiers. This fact may result in erroneously marking politicians' claims as unclear and evasive if our method is used in real-world scenarios without human monitoring, especially since the current state of LLMs under usage tends to hallucinate and produce unfaithful outputs. Hence, further work to ensure the reliability and trustworthiness of the underlying models would be crucial for their deployment.

\section*{Acknowledgments} The research work was supported by the Hellenic Foundation for Research and Innovation (HFRI) under the 3rd Call for HFRI PhD Fellowships (Fellowship Number 5537).

This work was supported by the Portuguese Recovery and Resilience Plan through project C64500888200000055 (NextGenAI - Center for Responsible AI), by the EU’s Horizon Europe Research and Innovation Actions (UTTER, contract 101070631), and by Fundação para a Ciência e Tecnologia through contract UIDB/50008/2020.

\bibliography{custom}

\clearpage
\appendix

\section{Dataset details}
\label{sec:dataset}

\subsection{Exploratory data analysis}
\label{sec:interviews}
\begin{figure}[h!]
    \centering
    \includegraphics[width=\columnwidth]{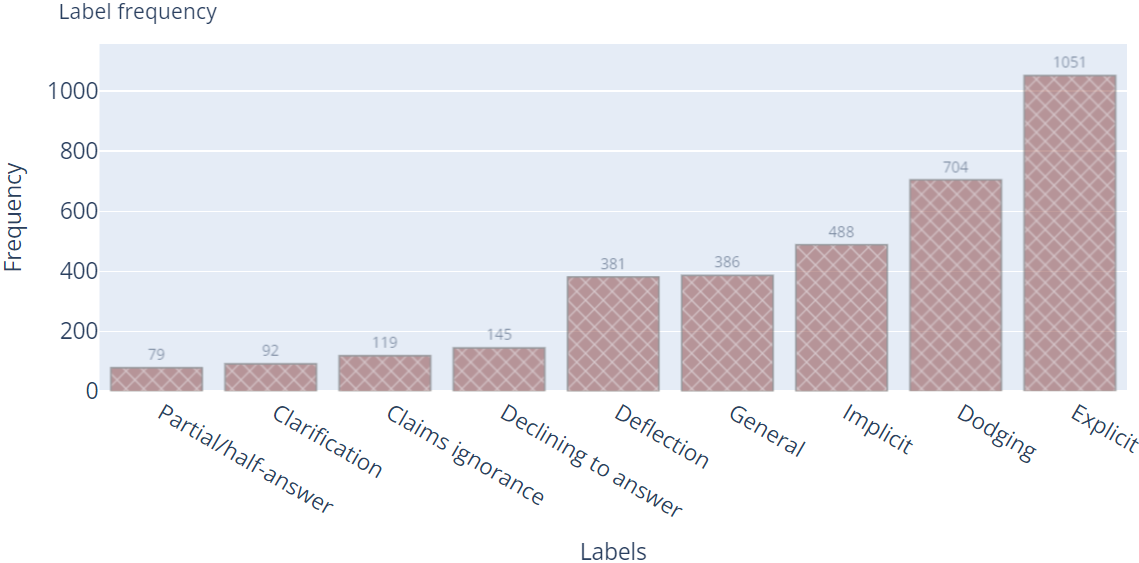}
    \caption{Label distribution in the dataset.}
    \label{fig:all-labels}
\end{figure}

In this section, we describe some interesting patterns present in our proposed dataset. 
\paragraph{Label distribution}
We start our analysis from the core of this work, which is the distribution of the final labels of our dataset, which are presented in Figure \ref{fig:all-labels}. Overall, \textit{Explicit Replies} is the most prevalent category, followed by evasion categories with significantly lower frequency each. Specifically, Explicit Replies contribute to 1051 samples in total, followed by Dodging (704 samples), Implicit (488 samples), General (386 samples), Deflection (381 samples), Declining to answer (145 samples), Claims ignorance (119 samples), Clarification (92 samples) and finally Partial/half-answer (79 samples).

\begin{figure}[h!]
    \centering
    \includegraphics[width=\columnwidth]{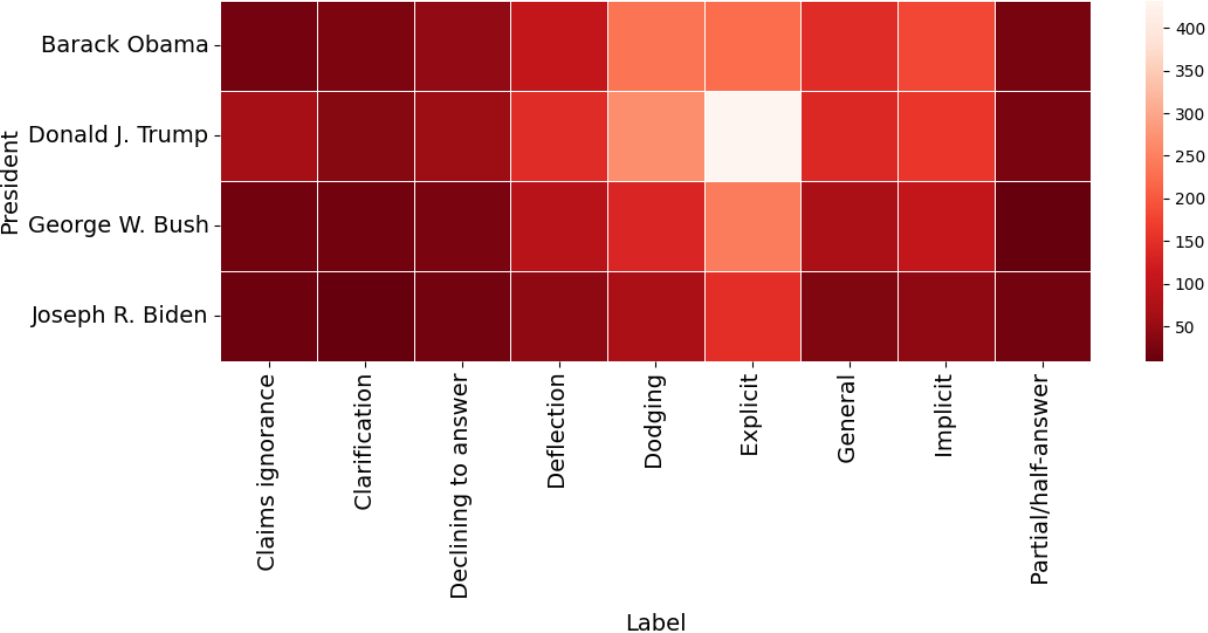}
    \caption{Label distribution per president.}
    \label{fig:president-labels}
\end{figure}

We also analyze the label distribution per president in Figure \ref{fig:president-labels}, offering a more detailed insight compared to Figure \ref{fig:barplot_pres}. According to the per president distribution, we conclude that in our collected interviews Donald J. Trump tends to provide more Explicit Replies than the rest of the US presidents, as indicated by the light-colored square of Figure  \ref{fig:president-labels}.

In the following paragraphs we will delve into the insights behind these label distributions.

\begin{figure}[h!]
\centering
    \includegraphics[width=0.5\textwidth]{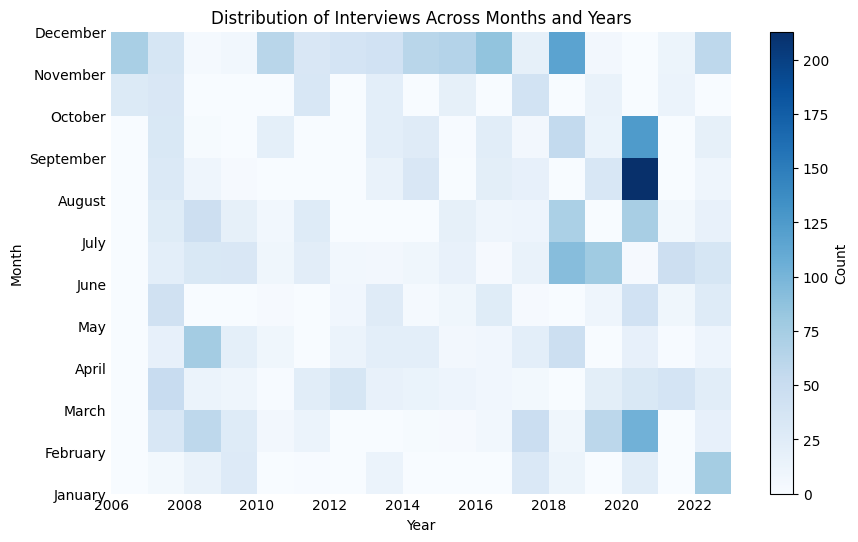}
\caption{Visualization of interview distribution across months and years in the corpus}
\label{fig:months_year_statistics}
\end{figure}

\paragraph{Temporal insights}
Moving on to temporal characteristics, in Figure \ref{fig:months_year_statistics} we provide some temporal statistics regarding the interview distribution.

\begin{figure}[h!]
\centering
    \includegraphics[width=\columnwidth]{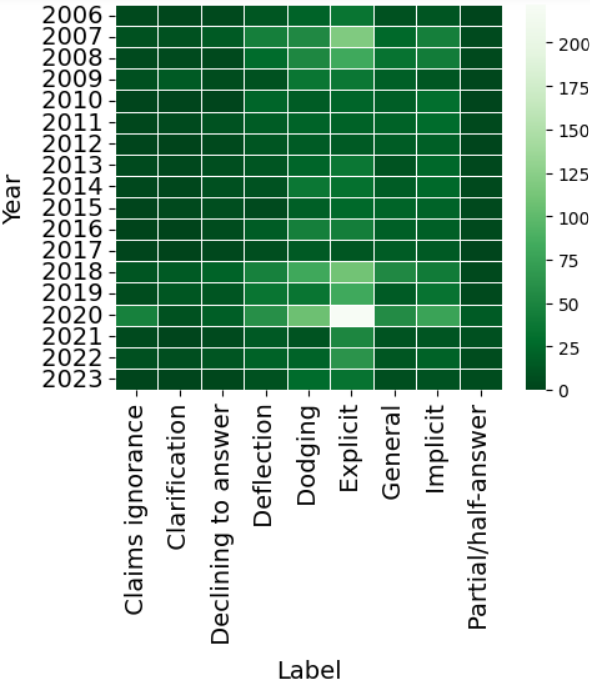}
\caption{Label distribution across years}
\label{fig:label-per-year}
\end{figure}

In Figure \ref{fig:label-per-year} we present the label distribution per year in our dataset. We observe an elevated number of Explicit Replies in 2020, as indicated by the light-colored cell. This observation can be grounded to president-related information, as this can be a strong characteristic in conjunction to label distribution.

\begin{figure}[h!]
    \centering
    \includegraphics[width=\columnwidth]{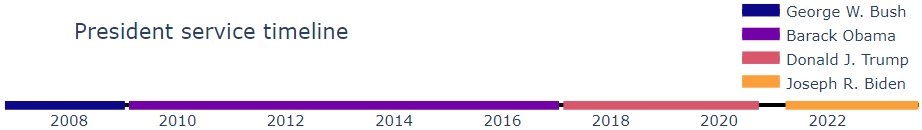}
    \caption{Service timeline for each US president}
    \label{fig:timeline}
\end{figure}

\begin{figure*}[h!]
    \hskip -0.21in \includegraphics[width=1.04\textwidth]{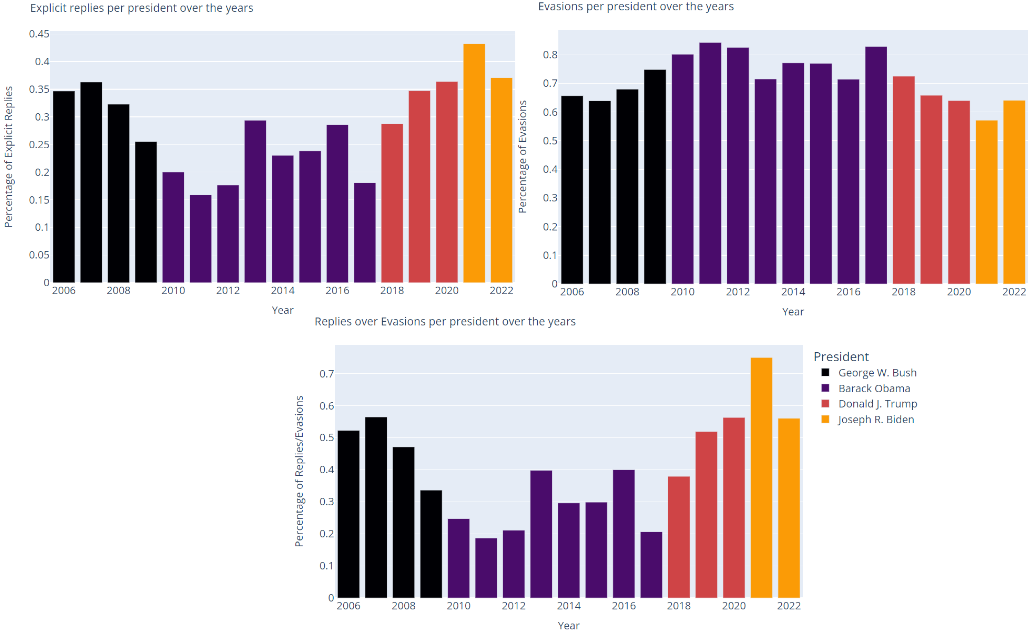}
    \caption{Percentages of Explicit Replies (left), Implicit/Non-Replies (right) and ratio of Replies over  Implicit/Non-Replies (bottom) for each US president during their service.}
    \label{fig:evasion}
\end{figure*}

So, in association with US presidents, in Figure \ref{fig:timeline} we demonstrate the timeframe associated with each president's service. We can now conclude that the higher number of Explicit Replies of Figure \ref{fig:label-per-year} coincides with Trump's service, which is related to more Explicit Replies, as indicated in Figure \ref{fig:president-labels}.
Consequently, temporal evasion characteristics are highlighted in Figure \ref{fig:evasion}.

\begin{figure*}[h!]
    \hskip -0.21in \includegraphics[width=1.04\textwidth]{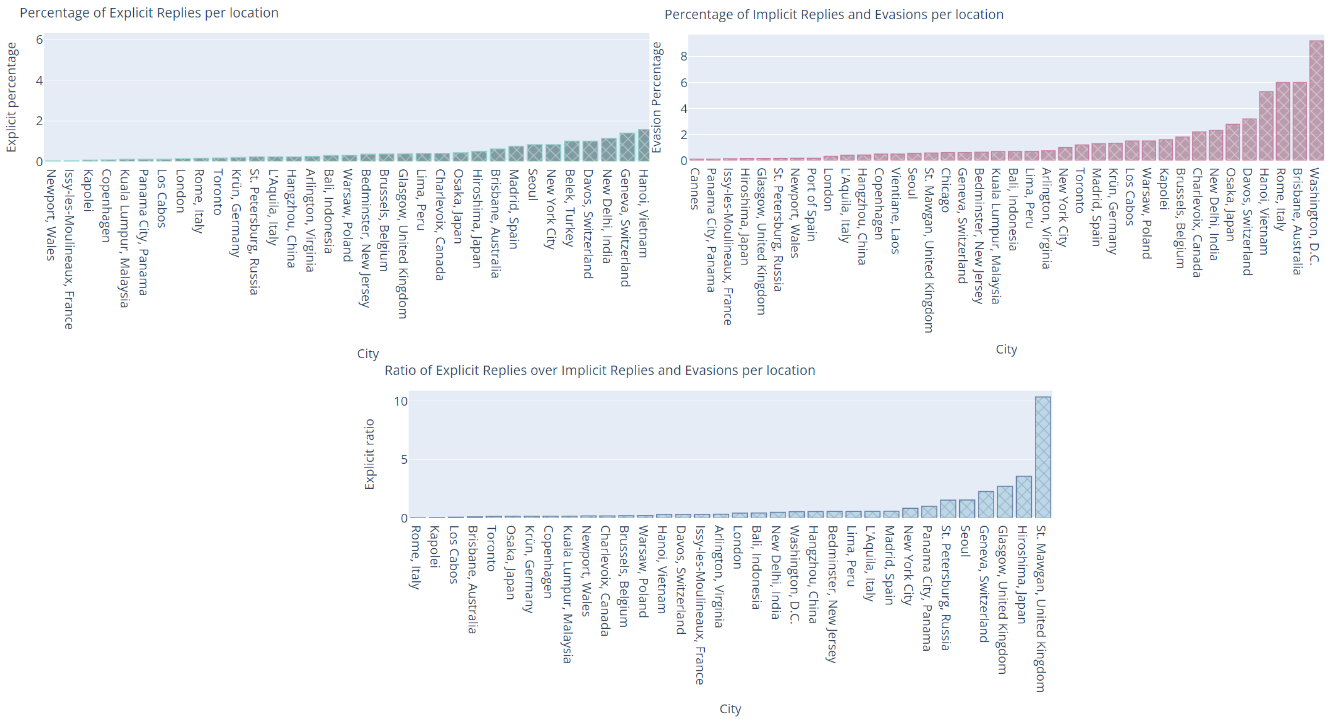}
    \caption{Percentages of Explicit Replies (left), Implicit/Non-Replies (right) and ratio of Replies over  Implicit/Non-Replies (bottom) per location.}
    \label{fig:location}
\end{figure*}

\begin{figure*}[h!]
    \centering
    \includegraphics[width=0.65\textwidth]{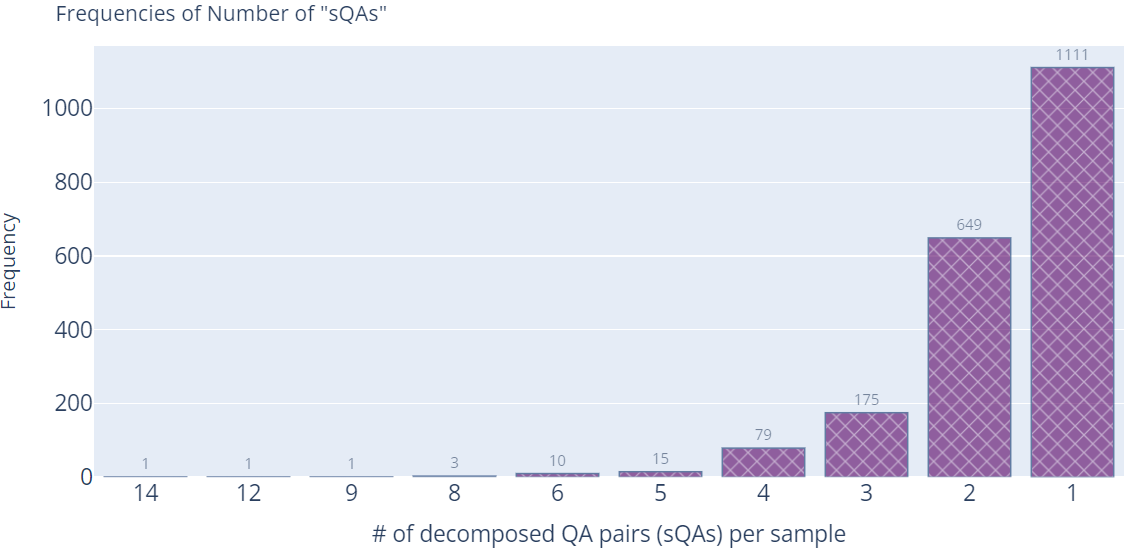}
    \caption{Distribution of sQAs length frequency.}
    \label{fig:summaries}
\end{figure*}

\begin{figure*}[h!]
    \centering
    \includegraphics[width=\textwidth]{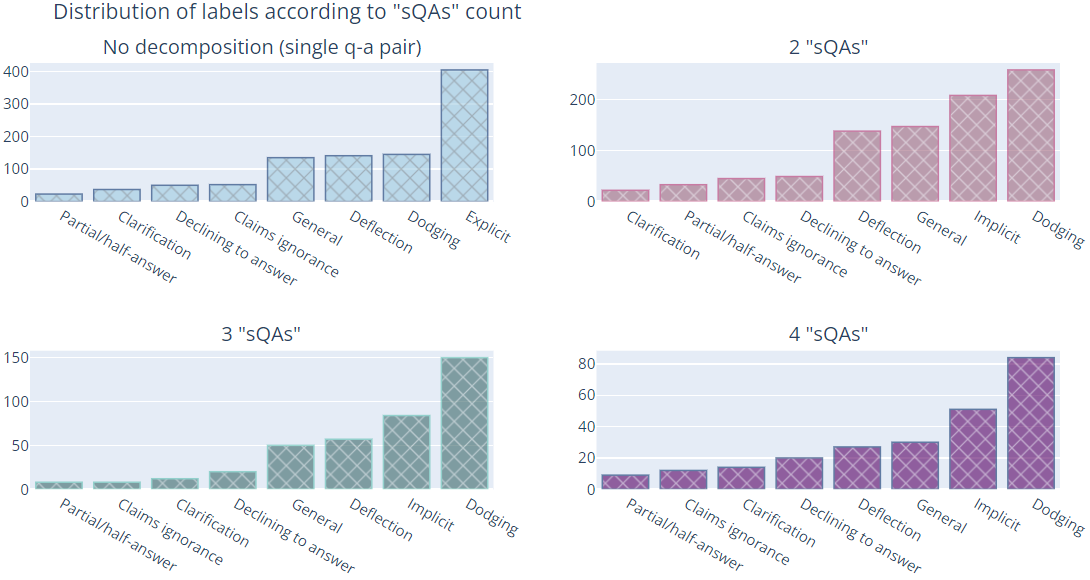}
    \caption{Label frequency per sQAs length.}
    \label{fig:summaries-labels}
\end{figure*}

\begin{figure}[h!]
    \includegraphics[width=0.5\textwidth]{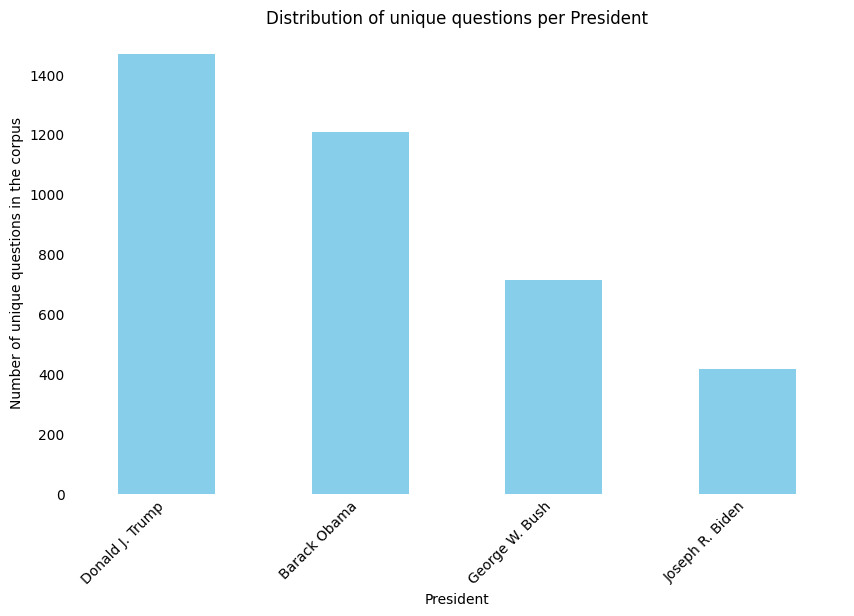}
\caption{Visualization of distribution of unique questions per President in the corpus}
\label{fig:presidents_distribution}
\end{figure}

\begin{figure*}[h!]
    \centering
    \includegraphics[width=\textwidth]{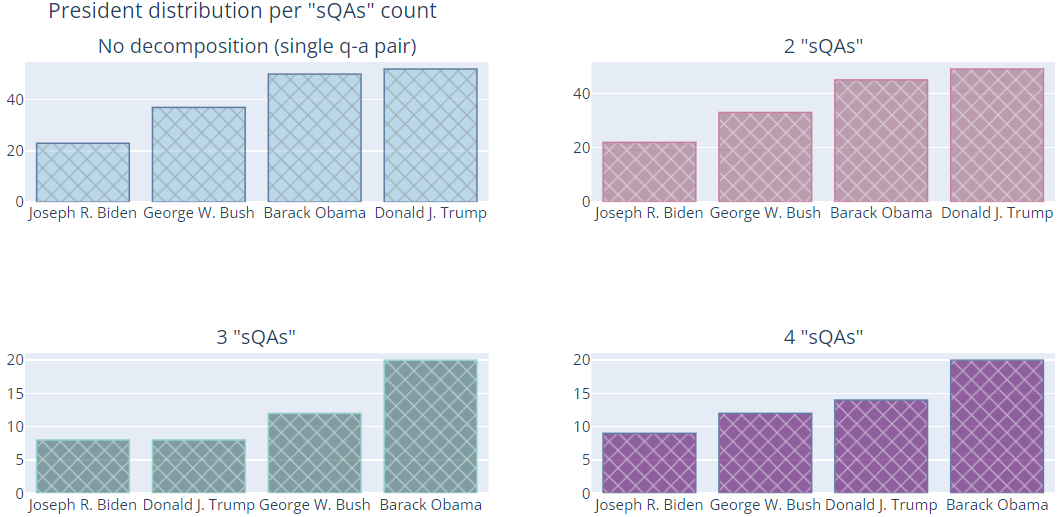}
    \caption{Distribution of per president interviews for different sQA counts.}
    \label{fig:president-summaries}
\end{figure*}

To this end, some interesting patterns can be derived from Figure \ref{fig:evasion}, especially if we focus on the start and the end of each president's service period. For example, George W. Bush and Joseph R. Biden tend to significantly decrease their ratio of Explicit Replies over implicit replies and evasion strategies, while the opposite pattern occurs for Donald J. Trump. Regarding Barack Obama, his ratio is almost the same at the end of his service in comparison to the beginning, even though fluctuations are observed during his entire service period.

\paragraph{Geographical insights}
Location-related patterns are examined in Figure \ref{fig:location} in order to derive whether evasion phenomena occur in conjunction to certain locations. Specifically, the horizontal axis represents the location where a presidential speech took place, while the vertical axis corresponds to the percentage of Clear Replies (left), Ambivalent Replies and Clear Non-Replies (right) and the ratio of these two cases (bottom). All percentages are normalized according to the total number of interviews given to each of those locations according to our data.
Focusing on the Explicit Reply ratio over all other cases (bottom plot), the resulting long-tailed distribution denotes that in most cases there are few Explicit Replies compared to evasion techniques or Implicit Replies.
Overall, we cannot extract a specific pattern location-wise, meaning that the evasion rate is not strongly associated with location.
\paragraph{QA decomposition}
We also analyze the distribution of sQAs, so that we discover the impact of the number of decomposed QA pairs on other dataset characteristics. This distribution is showcased in Figure \ref{fig:summaries}, where single QA instances dominate the dataset (the highest bar corresponds to 1 sQA, which is equivalent to the initial question and answer, and not decomposed by ChatGPT). As a general tendency, longer QAs -and therefore larger numbers of sQAs- are rare, as proven by the lower bars of Figure \ref{fig:summaries}. This observation eases the annotation process, since longer QA pairs are harder to decompose by ChatGPT, and are consequently evaluated and annotated by humans. 

An interesting insight that can be derived from the sQAs count per interview is the corresponding label distribution. This analysis is presented in Figure \ref{fig:summaries-labels} (we only consider the more frequently occurring sQA numbers as per Figure \ref{fig:summaries-labels}, i.e. instances with 2, 3, 4 sQAs or no sQA as in the case of non-decomposed QA pairs). Interestingly, the top-5 frequent categories are the same for sQAs of counts 2, 3, 4 (Dodging,  Implicit, General, Deflection, and Declining to answer categories). 
Moreover, Explicit Replies are absent from sQAs of count 2, 3, 4, even though they are frequent labels in the dataset (Figure \ref{fig:all-labels}).
This pattern differs for QA pairs with no decomposition (upper left plot): Explicit Replies are significantly more frequent, followed by other frequently occurring evasion categories (Deflection, General, Dodging). This analysis also suggests an important insight: politicians tend to provide clear replies in answers targeting short, single-barrelled questions while concealing evasion strategies within answers for multi-part questions, where grounding the requested information to the answer given is significantly harder.

Moving forward to a per-president analysis, details regarding the number of questions for all 4 US presidents existing in the interviews under consideration are provided in Figure \ref{fig:presidents_distribution}. 

We can then proceed by examining the per-president decomposition of questions. The related analysis is presented in Figure \ref{fig:president-summaries}. 

Barack Obama receives more multi-part questions, therefore scoring high in instances where there are 3 or 4 sQAs (bottom plots of Figure \ref{fig:president-summaries}). This can be possibly related to the elevated number of Ambivalent Replies and low number of Explicit Replies (Figure \ref{fig:barplot_pres}) in association with the connection between evasion frequency and number of sQAs per instance (Figure \ref{fig:summaries-labels}). On the other hand, Donald J. Trump scores higher in instances where single QA pairs occur, or are broken down into 2 parts (2 sQAs), as indicated by the top plots of Figure \ref{fig:president-summaries}. This could be related to the comparatively lower number of Donald J. Trump Ambivalent replies (Figure \ref{fig:barplot_pres}) and the higher number of Explicit Replies (Figure \ref{fig:president-labels}). 
\begin{figure*}[h!]
    \centering
\includegraphics[width=\textwidth]{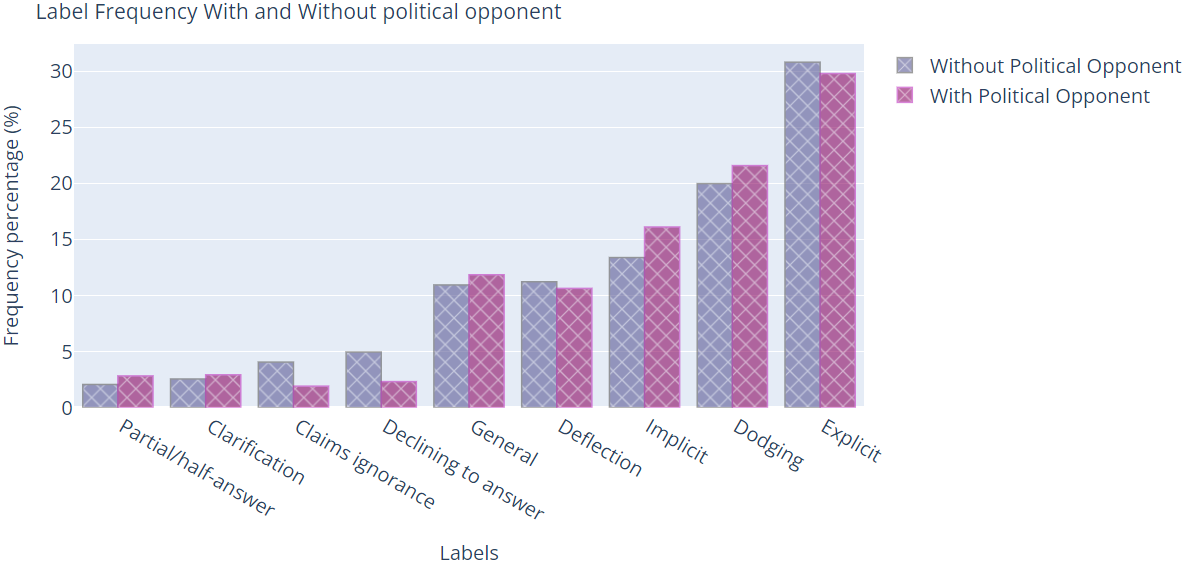}
    \caption{Label percentages for interviews with and without the presence of a political opponent.}
    \label{fig:opponent-label}
\end{figure*}

\begin{figure*}[h!]
 \centering
  \begin{subfigure}{0.495\textwidth}
    \includegraphics[width=1.02\linewidth]{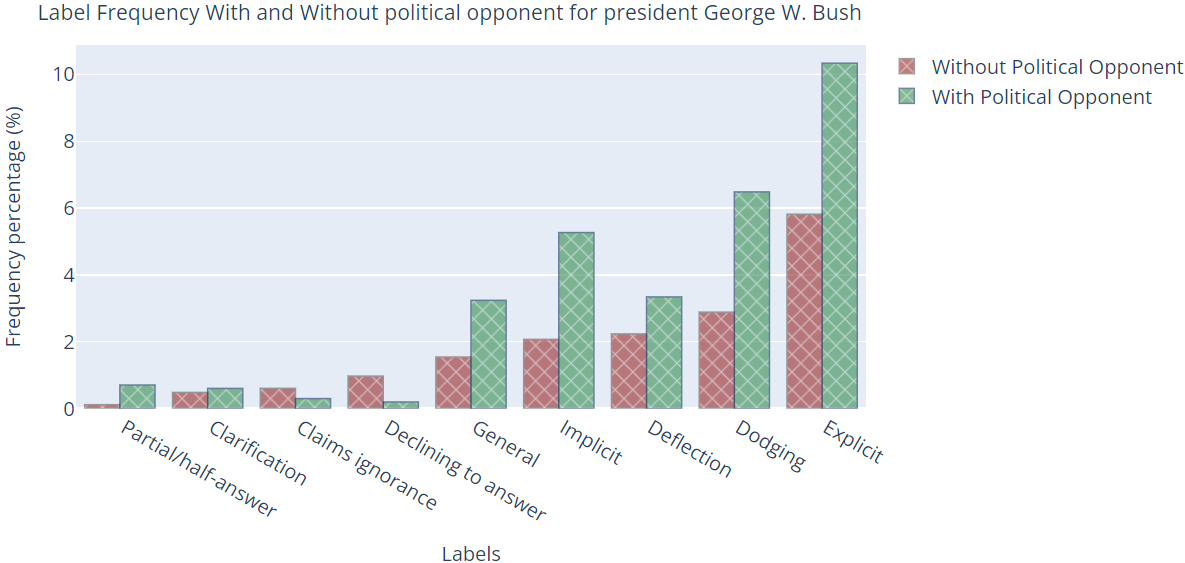}
    \caption{Label distribution for G. Bush.}
    \label{fig:bush}
  \end{subfigure}
  \hfill
  \begin{subfigure}{0.495\textwidth}
    \includegraphics[width=1.02\linewidth]{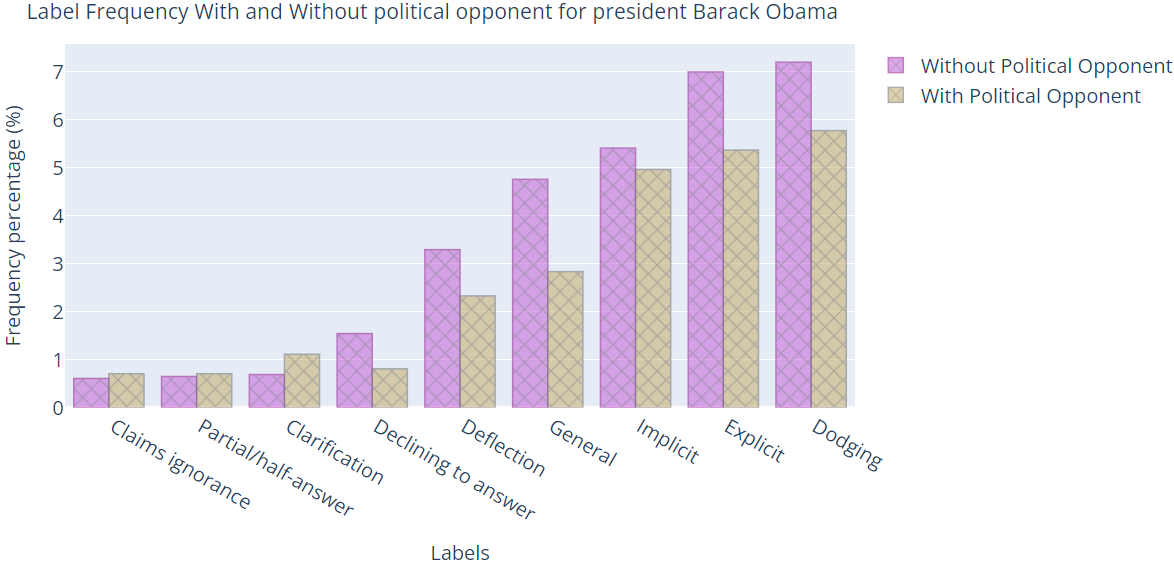}
    \caption{Label distribution for B. Obama.}
    \label{fig:obama}
  \end{subfigure}
  \medskip
  \vskip 0.05in
  \begin{subfigure}{0.495\textwidth}
    \includegraphics[width=1.02\linewidth]{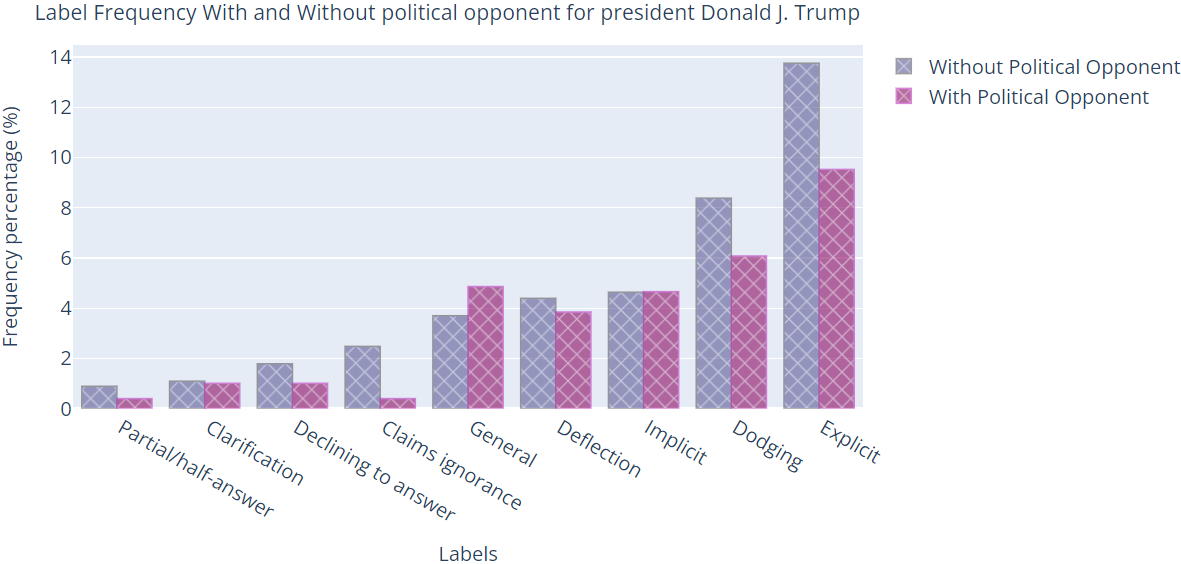}
    \caption{Label distribution for D. J. Trump.}
    \label{fig:trump}
  \end{subfigure}
  \hfill
  \begin{subfigure}{0.495\textwidth}
    \includegraphics[width=1.02\linewidth]{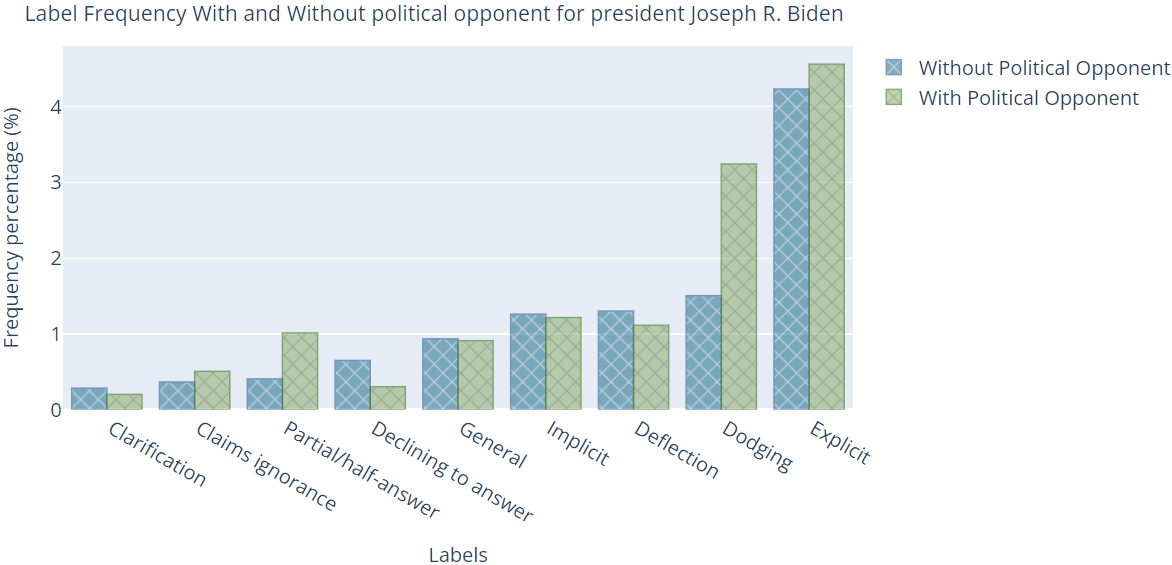}
    \caption{Label distribution for J. R. Biden.}
    \label{fig:biden}
  \end{subfigure}
  \caption{Label distribution with and without opponent for each US president of our dataset.}
  \label{fig:all-politicians-opponent}
\end{figure*}

To this end, our QA decomposition is deemed as an interesting initial tool towards the possibility of evasions: in cases where many multi-part questions occur, it is possible that evasion strategies may also appear, while the opposite holds in cases with single QA pairs.

\paragraph{Political opponents} In Figure \ref{fig:opponent-label} we present the distribution of labels when a politician is interviewed on their own versus when they are interviewed with a political opponent. Politicians are more or less consistent towards their Explicit Replies and evasion percentages, as proven by the similar bar height in both cases (with or without an opponent).

Delving deeper into the opponent-related analysis, in Figure \ref{fig:all-politicians-opponent}
we present label percentages with and without political opponent per president. Different patterns arise for each of them: for example, George Bush (Figure \ref{fig:bush}) tends to provide more Explicit Replies when being interviewed together with a component than when on his own.  
On the contrary, Barack Obama (Figure \ref{fig:obama}) provides more Explicit Replies when being interviewed on his own. Similarly, Donald J. Trump (Figure \ref{fig:trump}) replies explicitly when no opponent is participating in the interview.
Smaller differences in Explicit reply percentages under the two interview scenarios are observed for Joseph R. Biden(Figure \ref{fig:biden}), even though he tends to provide slightly more Explicit Replies in interviews with a political opponent. Donald J. Trump and Joseph R. Biden tend to employ evasion strategies in similar percentages with and without political opponents; some notable exceptions can be observed for Dodging categories, for which the percentages for Biden are higher in presence of a political opponent, while the opposite holds for Trump. In total, the label distributions for Barack Obama and Donald Trump are somewhat similar (note the ranking of labels, as well as the differences between bars with/without opponent), indicating a common behavior in handling interviews with/without political opponents. George Bush holds a diverging distribution, in terms of presenting a larger gap between his top-1 category (Explicit Replies) and the rest; especially when being interviewed on his own, he tends to exploit significantly less evasion techniques in comparison to the rest of the presidents.
\\ \\
Overall, our presented dataset accompanied by this exploratory analysis can be utilized by political scientists, assisting them in extracting interesting insights from political interviews.

\subsection{Examples from the proposed taxonomy}
\label{app:examples_taxonomy}
In Table \ref{tab:typology-examples}, we demonstrate some examples for all the categories mentioned in our proposed taxonomy. We also provide explanations on why these examples were classified in their respective classes.

\begin{table*}[h!]
\centering
\begin{tabular}{c| p{1.6cm} p{5.2cm} p{7.2cm}}
\toprule
\multicolumn{2}{c}{\small  \textbf{taxonomy}} & \small \textbf{Description} &  \small \textbf{Example} \\
\midrule
\multirow{3}{0.2cm}{\begin{turn}{90}
\small Clear R. \end{turn}} 
& \small Explicit & \small The information requested is explicitly stated (in the requested form) & \small \textbf{Q:} er you have your own views about PR at Westminster don’t you?
\textbf{A:} I do.

\textit{Why? - directly gives the info requested}\\
\midrule
\multirow{27}{0.2cm}{\begin{turn}{90}
\small Ambivalent Reply \end{turn}} 
& \small Implicit & \small The information requested is given, but without being explicitly stated (not
in the expected form) & \small \textbf{Q:} Are you going to watch television?
\textbf{A:} What else is there to do?

\textit{Why? - they suggest planning to watch TV, despite not explicitly stating it}\\
\cmidrule{2-4}
& \small General & \small The information provided is too general/lacks the requested specificity & \small \textbf{Q:} What's your favourite film?
\textbf{A:} Fight Club, Filth and Hereditary

\textit{Why? - the reply gives three movies instead of one, which makes the desired information unclear}\\
\cmidrule{2-4}
& \small Partial & \small Offers only a specific component of the requested information & \small \textbf{Q:} Did you enjoy the film?
\textbf{A:} The directing was great

\textit{Why? - Directing is only part of what constitutes a film}\\
\cmidrule{2-4}

& \small Dodging & \small Ignoring the question altogether & \small \textbf{Q:} Do you like my new dress?
\textbf{A:} We are late.

\textit{Why? - does not even acknowledge the question and goes straight to another topic}\\

\cmidrule{2-4}
& \small Deflection & \small Starts on topic but shifts the focus and makes a different point than
what is asked & \small \textbf{Q:} Did you eat the last piece of pie?
\textbf{A:} I have to admit that this was a great recipe, I always like it when there are chocolate
chips in the dough.

\textit{Why? - acknowledges the question but goes on a tangent about the chips, without answering}\\





\midrule

\multirow{13}{0.2cm}{\hspace{70cm}\begin{turn}{90}
\small Clear Non-Reply \end{turn}} 

& \small Declining to answer & \small Acknowledge the question but directly or indirectly refusing to
answer at the moment & \small \textbf{Q:} The hypothesis I was discussing, wouldn’t you regard that as a defeat?
\textbf{A:} I am not going to prophesy what will happen.

\textit{Why? - directly stating they won’t answer}\\
\cmidrule{2-4}
& \small Claims ignorance & \small The answerer claims/admits not to know the answer
themselves & \small \textbf{Q:} On what precise date did the government order the refit of the HMAS Kanimbla in preparation
for its forward deployment to a possible war against Iraq?
\textbf{A:} I do not know that date. I will find out and let the House know.

\textit{Why? - claims/admits they don’t have the information}\\
\cmidrule{2-4}
& \small Clarification & \small Does not provide the requested information and asks for clarification & \small \textbf{Q:} Was it your decision to release the fund?
\textbf{A:} You mean the public fund?

\textit{Why? - gives no data, asks for clarification}\\
\bottomrule
\end{tabular}
\caption{Descriptions and examples of political evasion techniques based on the proposed taxonomy}
\label{tab:typology-examples}
\end{table*}

These examples were used in the annotators' "training" phase, during which they were familiarized with the introduced problem, as well as the proposed taxonomy. The same examples were used as demonstrations for few-shot prompting, inserted in the same order as in Table \ref{tab:typology-examples}.
\begin{figure}[h!]
    \includegraphics[width=0.45\textwidth]{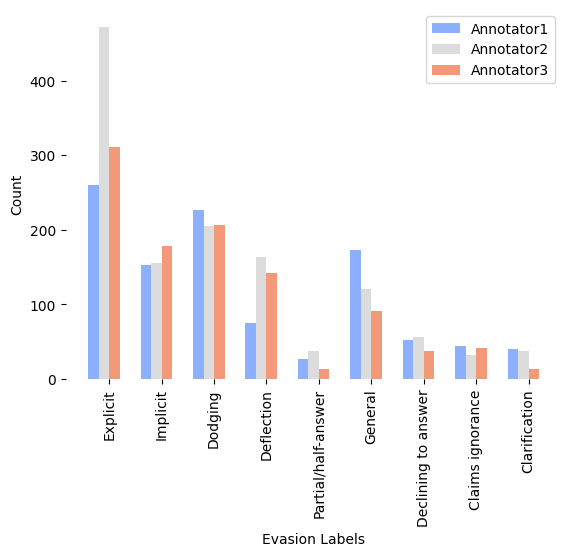}
\caption{Visualization of distribution of evasion label per annotator in the corpus}
\label{fig:annotators_stats}
\end{figure}
\subsection{Annotation details}
\paragraph{Annotators' statistics}
\label{sec:annotators}
All three non-expert annotators are of engineering background and participated in this annotation process voluntarily. The reason why we opted for non-expert annotators is because they are more representative of the general public, who are the receivers of political speech and do not have adequate background to immediately capture possible evasions, and therefore cannot fully evaluate the response clarity. The three non-experts are females, while the expert annotator is male, and all of them are fluent or native English speakers. We do not disclose geographical characteristics to fully preserve anonymity. Moreover, we did not collect any information regarding age or race/ethnicity.

\paragraph{Quality of annotations} was ensured via a well-crafted process of designing and monitoring the annotation process. First of all, we collect a descriptive set of instructions: as an introduction, we provided our annotators the examples of Table \ref{tab:typology-examples} to familiarize with the nature of the categories. Then, we released a short quiz to validate that they properly learned the fundamentals. After this stage, we proceeded with real examples from our dataset, demonstrating some examples of successful and unsuccessful sQAs in comparison to the initial interviews. Then, we also demonstrated examples with their labels to allow annotators to learn the distinguishing features between each category, especially the usually confused ones (as per Figure \ref{fig:heatmap}). Since this step is the most critical for the annotation process, we conducted daily sessions for one week, also distributing short quizzes after each session. The expert monitored and graded the learning process and the quizzes, verifying that the annotators were ready to perform annotations on their own, while also resolving any related questions in the meanwhile. Weekly checks on the annotation quality were performed by comparing a subset of the annotations with the annotations provided by the expert. In these intermediate evaluations, no annotator was significantly deviating from the expert. We denote that we consider a non-negligible deviation when the Fleiss score between the expert and any annotator was $\leq$ 0.7.

\paragraph{Label distribution per annotator} Figure \ref{fig:annotators_stats} depicts the distribution of evasion labels for each non-expert annotator (note that interview samples were randomly distributed to annotators). The analysis reveals a generally consistent number of labels for each category across annotators. Notably, a slight disparity is observed for the explicit label, with annotator2 exhibiting a significantly different count compared to the other annotators. However, it's important to note that this doesn't necessarily imply a higher likelihood of Annotator2 to annotate instances with this label, as such behavior is not evident in the broader dataset analysis. The observed variation may be attributed to factors such as differing annotation styles or a higher occurrence of explicit responses within Annotator2's set, which is in accordance to the higher number of explicit replies in general (Figure \ref{fig:all-labels}).
\begin{figure*}[h!]
    \centering
    \includegraphics[width=\textwidth]{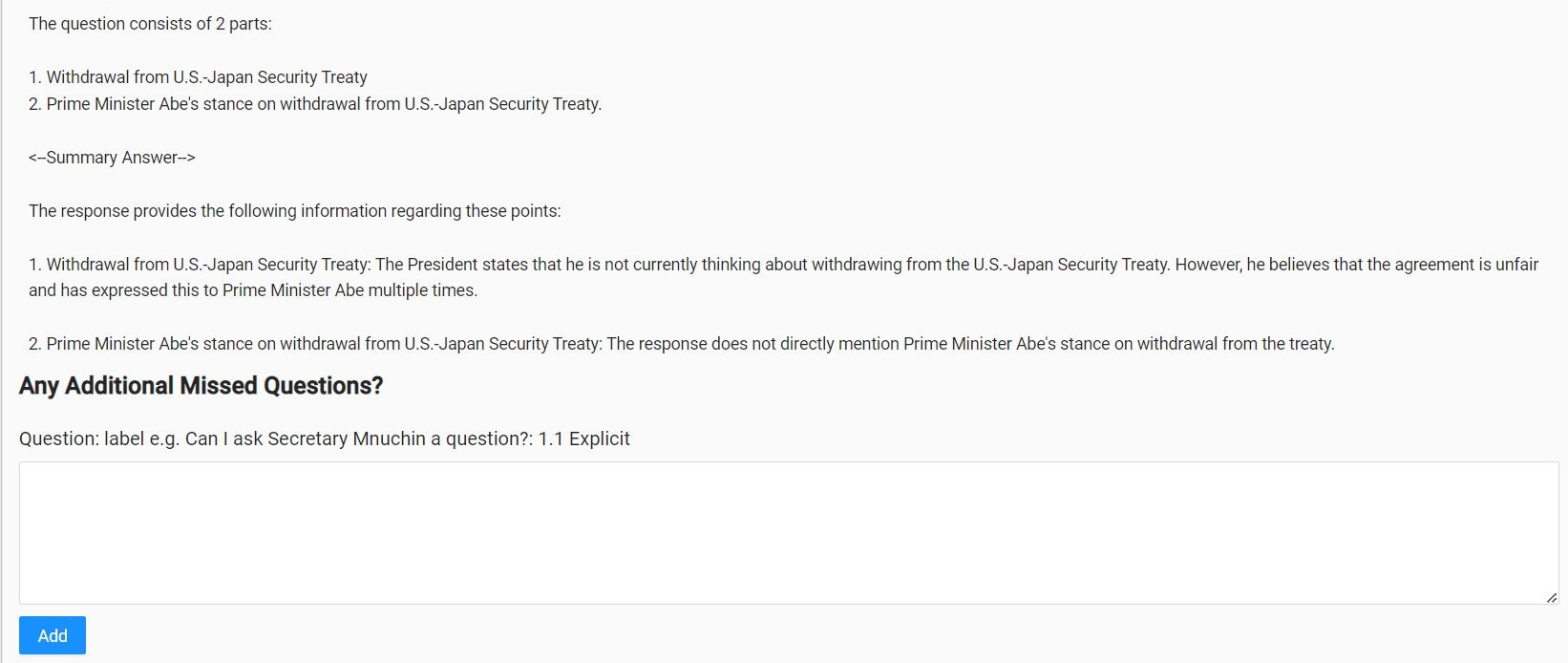}
    \caption{Screenshot from labelling platform: The sQAs for the provided QAs are given to the annotators. They have to highlight each of the enumerated responses and assign one of the labels of the taxonomy (as presented in Figure \ref{fig:label1}) to each of them.}
    \label{fig:label2}
\end{figure*}
\begin{figure*}[t!]
    \centering
    \includegraphics[width=\textwidth]{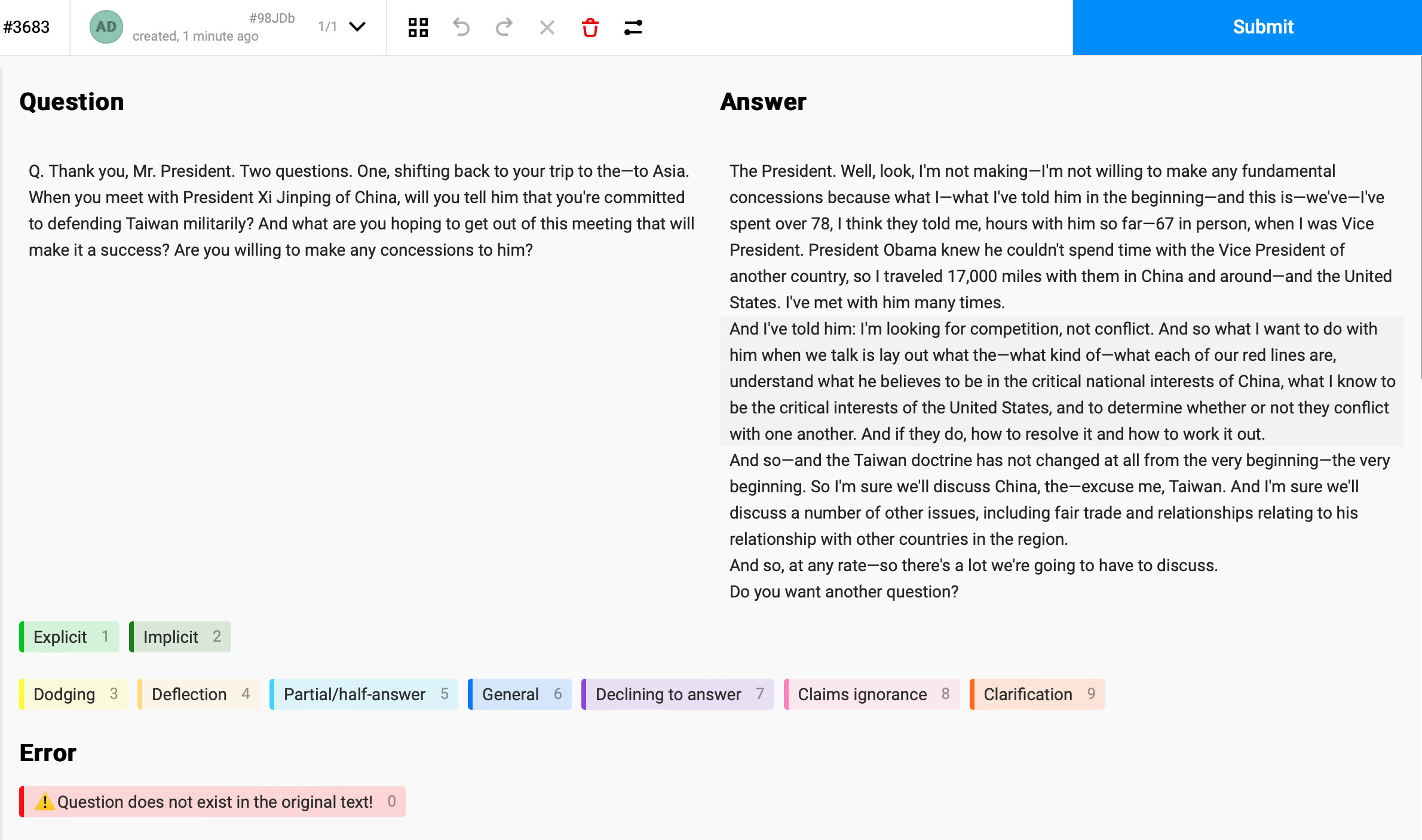}
    \caption{Screenshot from labelling platform: annotators have to read the original Question and Answer as provided. The classes corresponding to our proposed taxonomy are demonstrated as well.}
    \label{fig:label1}
\end{figure*}

\paragraph{Average annotation time per annotator} The average time taken by each annotator to complete the annotation of a segment of an interview was 144.33 seconds (2.4 minutes), excluding instances with exceptionally large durations. This metric directly reflects the inherent complexity of the annotation task. Notably, this average annotation time remained consistent across all annotators.

\paragraph{Labelling platform}
Our labelling process was conducted in the open source Label Studio\footnote{\href{https://labelstud.io/}{https://labelstud.io/}} platform. We provide some screenshots of the labelling pages in Figures \ref{fig:label2}, \ref{fig:label1}  (they both belong to the same labelling page). Before the labelling process commenced, we provided detailed guidance to annotators on how to use the platform properly, so that any erroneous annotations because of limited familiarization with the platform are eliminated.

Annotators have to first evaluate the decomposition quality of sQAs (Figure \ref{fig:label2}) as provided by ChatGPT. In case of erroneous decomposition, they have to add the corresponding multi-parts missing (``Any Additional Missed Questions$?$``), among with their taxonomy label. If extraneous multi-parts are generated by ChatGPT, they can be reported (annotators can click the Error button denoting that ``Question does not exist in the original text$!$``), so that this multi-part pair is disabled from the annotation process.

\paragraph{Annotations on presidential speech}
Extending the findings presented in Figure \ref{fig:barplot_pres}, Table \ref{tab:demo} demonstrates more thorough results regarding the clarity of responses, as well as the evasion schemas leveraged by US politicians, as a result of our annotations. All of them tend to provide Ambivalent Replies more often than not, as denoted with \textcolor{red}{red color}. Especially Barack Obama utilizes Ambivalent responses more frequently than the rest of the presidents. \textcolor{blue}{Blue color} denotes the most frequently used evasion technique, which in this case corresponds to `Explicit Replies`; nevertheless, Explicit Replies only account for about the 1/3rd of the responses for all presidents, leaving much space for evasion schemas to appear. In comparison, Joe Biden tends to provide more Explicit Replies, as resulting from our annotations.
\begin{table}[h!]
\small
\begin{tabular}{cP{0.7cm}P{0.7cm}P{0.7cm}P{0.7cm}}
\toprule
Response  & G. W. Bush & B. Obama & D. J. Trump & J. R. Biden \\
\midrule
 Clear Reply & 34.31 & 22.38 & 32.6 & 37.34    \\ 
 Clear Non-Reply & 8.68 & 9.5 & 11.77 & 10.53    \\ 
 Ambivalent & \textcolor{red}{57.0}  & \textbf{\textcolor{red}{68.12}}  & \textcolor{red}{55.62}  & \textcolor{red}{52.13}   \\
 \midrule
Explicit & \textcolor{blue}{34.31} & \textcolor{blue}{22.38} & \textcolor{blue}{32.6} & \textbf{\textcolor{blue}{37.34}}  \\
Implicit & 14.43 & 18.02 & 12.08 & 10.78 \\ 
Dodging & 19.05 & 23.17 & 20.08 & 17.54  \\
Deflection & 12.32 & 10.3 & 11.02 & 10.78  \\
Partial/half-answer & 1.4 & 2.28 & 1.96 & 5.01  \\
General & 9.8 & 14.36 & 10.49 & 8.02 \\ 
Declining to answer & 3.64 & 4.55 & 4.08 & 4.76 \\ 
Claims ignorance & 2.52 & 2.18 & 4.91 & 3.51  \\
Clarification & 2.52 & 2.77 & 2.79 & 2.26  \\
\bottomrule
\end{tabular}
\caption{Statistics of answer clarity and evasion techniques in political interviews per president.}
\label{tab:demo}
\end{table}

\paragraph{Dialogue separation} The annotators were tasked with identifying potential errors generated by ChatGPT. In Figure \ref{fig:label2}, they were presented with the option: `Error, Question does not exist in the original text.' Additionally, if any multi-part pairs were missing, annotators were encouraged to provide them, as shown in Figure \ref{fig:label1} with the prompt `Any additional missing questions?'
During the analysis of dialogue separation performed by GPT-3.5-turbo, it was found that 88.6\% of the segmented sections were accurately separated, with no errors detected in the sub-questions within the two incorrect segments. Conversely, only 11.4\% of the segments contained at least one error in the dialogue separation process. Specifically, 91.41\% of the sub-questions were deemed accurate, 7.31\% were labelled as `Error, Question does not exist in the original text,' and 1.27\% were initially missing questions that were later provided by the annotators.

\paragraph{Counterfactual Singular QAs (sQAs)} \label{sec:countersummaries}Considering that annotators should consult the initial interview text instead of exclusively relying on the more easily readable QA ChatGPT sQAs, we test their cautiousness by inserting 31 additional samples containing counterfactual sQAs in place of the original ones --without them knowing. Those sQAs are purposely unfaithful to the original QAs, guiding an annotator towards believing the responses belong to a different category compared to the actual one. We prompt ChatGPT to select an incorrect (counterfactual) label in order to generate a suitable sQA, which is shown to users instead of the original (the class label is not shown). \footnote{We provide the counterfactual sQA prompt at §\ref{sec:prompting}} We manually verify the suitability of each counterfactual sQA. The sQA should be marked as erroneous, and the annotator should write down the decomposed answers occurring, together with their labels.
\paragraph{SQAs insights} We computed for each annotator the ratio of selecting the counterfactual label instead of the correct one and found it to be $\leq0.08$. We thus assert that annotators do not solely rely on ChatGPT sQAs and confirm the validity of the process, since they were not significantly influenced by the counterfactual sQAs.

\section{Experimental Details}
\label{sec:exp-details}

In our experiments, we utilized three distinct datasets: training, development, and validation sets. The original dataset was divided into two parts, allocating 2700 samples to the training set and reserving approximately 750 samples for the development set. For a realistic evaluation, we employed a separate validation dataset comprising 274 samples, which were meticulously annotated by a team of annotators. Any inconsistencies were resolved by a domain expert. This method ensures a robust assessment of the models using ground truth labels validated by an expert. The distribution of each category across these datasets is depicted in Table \ref{tab:clarity_frequency} for clarity labels and Table \ref{tab:evasion_frequency} for evasion labels.

\begin{table}[h]
\small
\centering
\begin{tabular}{c|ccc}
\toprule
Label              & Train & Development & Validation \\ \midrule
Clear Reply 	& 796 &	 255 &	 86 \\
Ambivalent Reply &	 1617 &	 421 &	 207 \\ 
Clear Non-Reply &	 284 	& 72 &	 24 \\ 
\bottomrule
\end{tabular}
\caption{Distribution of Instances Across Clarity Labels in Training, Development, and Validation Sets.}
\label{tab:clarity_frequency}
\end{table}

\begin{table}[h]
\small
\centering
\begin{tabular}{c|ccc}
\toprule
Label              & Train & Validation & Test \\ \midrule
Explicit           & 796   & 255        & 90   \\ \hline
Implicit           & 381   & 107        & 59   \\ 
General            & 313   & 73         & 50   \\ 
Partial/half-answer& 69    & 10         & 3    \\ 
Dodging            & 563   & 141        & 61   \\ 
Deflection         & 291   & 90         & 27   \\ \hline
Clarification      & 69    & 23         & 4    \\ 
Declining to answer& 117   & 28         & 11   \\ 
Claims ignorance   & 98    & 21         & 10   \\ 
\bottomrule
\end{tabular}
\caption{Distribution of Instances Across Evasion Labels in Training, Validation, and Testing Sets.}
\label{tab:evasion_frequency}
\end{table}

\subsection{Evaluation}
\label{sec:eval}
Throughout our paper, we utilize classification metrics for evaluation. Specifically, accuracy, precision, and recall are employed, as well as F1 scores. Regarding F1, we use both the macro and the weighted average strategies. The macro F1 score is calculated as the average of the F1 scores for each class (see Eq. \ref{eq:f1_macro}), without considering the class distribution, whereas the weighted F1 score accounts for class frequency, giving more weight to larger classes (see Eq. \ref{eq:f1_weighted}).

\begin{equation}
F1_{macro} = \frac{1}{N} \sum_{i=1}^N F1_i
\label{eq:f1_macro}
\end{equation}

\begin{equation}
    F1_{weighted} = \sum_{i=1}^N \left(\frac{n_i}{N} \times F1_i\right) 
    \label{eq:f1_weighted},
\end{equation}

where \( n_i \) is the number of instances in each class.

\section{Performance Analysis for Each Class}
\label{app:per_per_class}

In this section, the performance of the instruction-tuned models, which have shown the best performance compared to other strategies, is presented by class. Table \ref{tab:weighted_performance} illustrates the performance of these models using a weighted strategy.

\begin{table}[tb!]
\centering
\small
\begin{tabular}{P{1.2cm}P{1.5cm}P{0.5cm}P{0.5cm}P{0.5cm}P{0.5cm}}
\toprule
\small Classification variant & \small Model  & \small Acc. & \small Prec. & \small Recall & \small F1    \\ \midrule
\multirow{5}{4em}{\small direct clarity} 
&\small Llama-7b  &  0.489	& 0.581	& 0.489	& 0.504 \\ 
&\small Llama-13b  & 0.587	& 0.719	& 0.587	& 0.594 \\ 
&\small Llama-70b  & \textbf{0.75}	& \textbf{0.757}	& \textbf{0.75}	& \textbf{0.75}2 \\ 
&\small Falcon-7b  &   0.294	& 0.537	& 0.294	& 0.186\\
&\small Falcon-40b   & 0.341	& 0.656	& 0.341	& 0.244\\ 
\midrule
\multirow{5}{4em}{\small evasion-based clarity} & 
\small Llama-7b & 0.662	& 0.669	& 0.662	& 0.665\\
& \small Llama-13b & 0.675	& 0.68	& 0.675	& 0.677 \\
& \small Llama-70b & \underline{0.713}	& \underline{0.743}	& \underline{0.713}	& \underline{0.72} \\
&\small Falcon-7b  & 0.533 & 	0.537	 & 0.533	 & 0.533 \\
&\small Falcon-40b  & 0.618	& 0.633	& 0.618	& 0.622 \\
 \bottomrule
\end{tabular}
\caption{Classification results using a weighted strategy, which averages F1 scores, weighted by class size. The best results for each strategy are \underline{underlined} and the best results overall are also in \textbf{bold}.}
\label{tab:weighted_performance}
\end{table}

Using the weighted strategy, the conclusions remain the same, although the numerical results are slightly improved. Further analysis of the model's performance for each class can be found in Table \ref{tab:perf_per_class_clarity}, which showcases the classification report of the tuned Llama-2-70b model with evasion-based clarity for each class, which has shown the best results among the other strategies.

\begin{table}[h]
\small
\centering
\begin{tabular}{c|cccc}
\toprule
 & Prec.  &  Recall &  F1 &   Sup. \\ \midrule
 Clear Reply &      0.54   &   0.74   &   0.62     &   84 \\
 Ambivalent  &     0.84   &   0.71   &   0.77     &  208 \\
Clear Non-Reply  &     0.63   &   0.68   &   0.65     &   25 \\
        
\midrule
        Acc.  &            &          &   0.71     &  317 \\
       Macro avg  &     0.67   &   0.71   &   0.68     &  317 \\
    Weighted avg  &     0.74   &   0.71   &   0.72     &  317 \\
\bottomrule
\end{tabular}
\caption{Classification report of the tuned Llama-2-70b model, for each class, demonstrating precision, recall, F1 score, and support.}
\label{tab:perf_per_class_clarity}
\end{table}

Notably, the model demonstrates its highest precision with the Ambivalent category at suggesting strong accuracy in identifying relevant instances, albeit with a moderate recall. This is followed by a decent performance in the Clear Non-Reply category, with a balanced precision and recall. The category Clear Reply, while having a high recall, indicating effective identification of most relevant cases, shows the lowest precision, which may indicate a higher rate of false positives. This issue particularly arises from confusion between Clear Replies and Ambiguous responses, and between Clear and General responses, as further analyzed in App. \ref{sec:evasion-classification}.

Overall, the model achieves a general accuracy of and similarly balanced macro and weighted average scores. These results indicate a reasonably good model performance, particularly in distinguishing the more frequently occurring Ambivalent category.

\section{Additional Experiments}

\subsection{Few-Shot prompting}
\label{sec:few-shot}
In the few-shot setup, we showcase the model results irrespective of their size. Unlike in the ZS setup, smaller models demonstrated better adherence to the output template and exhibited fewer hallucinations overall. Since the examples in our dataset are quite lengthy, we opt to select one example for each label to present to the model, along with the corresponding explanation provided in Table \ref{tab:llm-results-fs}. This methodology mirrors what the human annotators saw before commencing the annotation procedure. We noticed that Falcon struggled more to respond within the given template compared to the zero-shot approach. Nevertheless, examples in the few-shot setup seemed to aid the Llama-70b model in understanding the task, along with the smaller models. In the FS setup, the Llama-7b model exhibited comparable results to a model ten times larger in the ZS setup. 
In evasion-based clarity models, examples in the middle are often ignored. Instead, responses tend to align with the labels of the first or last examples. This phenomenon is well-documented in literature \cite{dong2022survey}. For example in Llamma-70b, 60\% of responses matched the labels of the final four examples, compared to less than 10\% in the ground truth.
\input{tables/fewshot}

\subsection{Answer Grounding}
\label{sec:ans_ground}

In this section, we outline the distinctions in model performance between single and multi-part questions. Specifically, we divided the test set into two distinct parts: one consisting of segments of the interview containing only single questions (112 out of 317 questions), and the other containing only segments with multi-part questions (205 out of 317 questions). We then compared the performance of each method. Using this methodology, we discovered that regardless of the method employed, every model exhibited lower performance on multi-part questions compared to single ones. The results for instruction-tuned models are shown in Table \ref{tab:loras_single}, while those for the prompting techniques applied to the model with the best results are presented in Table \ref{tab:gpt_single}. For each model or method, there are two lines: the first represents performance on the multi-part question set, and the second represents performance on the single question set.

To further investigate whether this difficulty is also encountered by humans, we compared the Fleiss score of the annotators between these two subsets. We found that the difference was only 0.03, indicating that there was no significant difference in the performance of annotators between single and multi-part questions. This suggests that the challenge of grounding answers to multi-part questions is unique to LLMs.

\input{tables/multi_questions_lora}

\input{tables/multi_questions_gpt}

\subsection{Connection to encoded knowledge}
\label{sec:prior_knowledge}

We further delve into the integral relationship between clarity classification and the knowledge pertaining to a specific named entity. Named entities frequently have properties that are considered common knowledge and that is why they are not explicitly mentioned in a response. As a result, the systems that try to define the clarity of a response would need to be aware of these properties of the name entities. In our dataset the most occurring named entities are persons' names, that why we focused the experimental analysis on these terms. Specifically, we split our dataset into two distinct parts, one containing only parts of the interview that include at least one person's name either in the interview question or the answer and a second one which contains no person names. The first set consists of 189 questions and the second of 128 questions. The differences between the performances for instruction-tuned models are shown in Table \ref{tab:ner_lora}, while those for the prompting techniques applied to the model with the best results are presented in Table \ref{tab:ner_gpt}. 

The results show that across all models and methods, the performance on the set without named entities is increased compared with the performance on the set with named entities. Notably, there was a steep improvement in the smaller, less knowledgeable models compared to the others, corroborating the findings of \cite{sun2023head}. In this case, if we apply the same comparison for the human-curated annotations, we can see that there was a difference of 0.1 in Fleiss score between the two subsets, implying that it was slightly more difficult for humans also to annotate the set with named entities compared to the other one. 

\input{tables/knowledgeability}
\input{tables/knowledgeability_gpt}

\section{Evasion classification}
\label{sec:evasion-classification}
In this section, we present the results of the evasion (low-level) classification problem. Table \ref{tab:lora_evasion_classification} illustrates the performance of the instruction-tuned model on the evasion classification problem, while Table \ref{tab:gpt_evasion_classification} showcases the performance using zero-shot and chain-of-thought prompting on the ChatGPT which is the best-performing model. The performance of the models on the evasion classification task is lower compared to the clarity classification. Among the instruction-tuned models, Llama-70b exhibits the best performance across all metrics, similar to the evasion classification model.

In ChatGPT, a higher level of performance is observed in the zero-shot setup compared to the chain-of-thought (CoT) for evasion classification, contrary to the evasion-based classification method. Further investigation reveals that employing CoT ChatGPT leads to greater confusion between the classes \textit{General} and \textit{Implicit}, as well as \textit{Implicit} and \textit{Partial/half-answer}, compared to the zero-shot setup, where the primary confusion lies between \textit{Partial/half-answer} and \textit{Explicit}. However, the confusion stemming from the zero-shot setup results in different clarity labels, unlike CoT, which elucidates the performance disparity between the two tasks. It is noteworthy that the challenge of discriminating between these classes persists even for humans, as evidenced by the lowest agreement between annotators for these labels, as indicated in Figure \ref{fig:heatmap}. This underscores a general difficulty in distinguishing between these two evasion strategies. This analysis is particularly intriguing, especially given the context where the model has not been exposed to the annotated data of the users.

\input{tables/evasion_classification}

In order to evaluate the performance of the models at the evasion level, Table \ref{tab:perf_per_class_clarity} displays the classification report of the best performing model, Llama-70b.

\begin{table}[h]
\small
\centering
\begin{tabular}{c|cccc}
\toprule
 & Prec.  &  Recall &  F1 &   Sup. \\ \midrule
Explicit   &    0.68   &   0.84  &    0.75     &   94 \\
Implicit  &     0.50  &    0.29  &    0.36     &   64 \\
Dodging   &    0.53   &   0.68   &   0.59     &   60 \\
Deflection   &    0.33  &    0.45   &   0.38   &     20 \\
Partial/half-answer    &   0.00  &    0.00   &   0.00     &    6 \\
General   &    0.55   &   0.37   &   0.44    &    49 \\
Declining to answer  &     0.46  &    0.60  &    0.52   &     10 \\
Claims ignorance  &     0.67  &    0.80   &   0.73     &   10 \\
Clarification    &   1.00   &   0.50    &  0.67   &      4 \\
\midrule
 Acc. &           & &                0.57  &     317        \\
              Macro avg    &   0.57    &  0.50   &   0.51   &    317  \\
           Weighted avg    &   0.56   &   0.57   &   0.55   &    317  \\
\bottomrule
\end{tabular}
\caption{Classification report of the tuned Llama-2-70b model, for each class, demonstrating precision, recall, F1 score, and support.}
\label{tab:perf_per_class_clarity}
\end{table}

The results indicate varying performance across different response types in the model's classification capabilities. For example, the ``Explicit'' category shows strong performance, resulting in a relatively high F1-score, which suggests the model is quite effective at identifying and correctly classifying explicit responses. In contrast, the ``Implicit'' and ``Deflection'' categories exhibit lower precision and recall, indicating challenges in accurately detecting and classifying these subtler forms of responses, similar to human annotators, as depicted in Table \ref{fig:heatmap}. Notably, the ``Clarification'' category achieved perfect precision but lower recall, highlighting that while the model is highly accurate when it identifies these responses, it consistently fails to detect them.

\section{Encoder models}
\label{sec:encoder-classification}
In this section, to evaluate the performance of smaller models on the proposed task, we trained three different architectures: DeBERTa \cite{he2021deberta}, RoBERTa \cite{liu2019roberta}, and XLNet \cite{yang2019xlnet}, and assessed their performance on the same test set. Specifically, we selected two different sizes for each model: base and large, to examine the impact of size variation on model performance. The primary challenge we encountered was truncation, as the maximum input size for DeBERTa and RoBERTa is 512 tokens. To ensure a fair comparison, we also utilized XLNet, which does not have inherent input size limits. We fine-tuned these models using only non-truncated inputs to reduce noise during training. Specifically, out of the total 2700 samples in the training set, only 1713 (63\%) had fewer than 512 tokens. We trained the models for five epochs with a constant learning rate of $10^{-5}$. Evaluation of the models was conducted using the same test set, without removing 173 out of 317 samples with more than 512 tokens. The evaluation results are presented in Table \ref{tab:encoder-results}, while Table \ref{tab:encoder-not-truncated-results} displays the results of the same models on the subset of 173 samples with non-truncated inputs. For comparison, the results of the instruction-tuned LLama models on this subset are also included. As shown in Table \ref{tab:encoder-truncated-results}, the performance of the models on the subset with truncated inputs is close to random chance.

Another noteworthy finding is that the base models consistently outperformed their respective larger counterparts. Specifically, the output of every large model collapsed to a single label. For instance, RoBERTa-large with evasion-based clarity returned the label "Explicit" for every sample. Similar behaviour was observed for every large variant of the three different models.

To further evaluate the behaviour of encoder models and to explain their performance, we again check the differences in performance between the set of entities with named entities and without. The results are shown in Table \ref{tab:ner_enc}. The first line of each model displays the results for the set containing only interview parts with named entities, while the second line shows the results for the parts without named entities. The `large' variations of the models were omitted as they returned only a single class regardless of their input. This shows that the performance of encoders in the subset without named entities was improved for every model, regarding the classification strategy. Again, we evaluate the performance of the encoder in the subset and single-part questions, and the results are depicted in Table \ref{tab:single_enc}. The results show that the performance of the models in the subset that contains multipart questions is near to random chance, probably due to increased input size which increases the probability of truncation. This behaviour is consistent even for the XLNet model, where there is no length restriction in their input, so truncation does not occur. However, an interesting observation is that for single-part questions, the models, especially RoBERTa and XLNet, have comparable performance with generative models such as Llama-70b.
\input{tables/encoders_ins}

\input{tables/encoders_results}

\section{Comparison with Relevant Tasks}
\label{app:comparison_intent}
In this section, we compare the focus of our work to the closely related work of \citet{ferracane-etal-2021-answer}. The relevance of this analysis stems from the general similarity between our analysis and theirs, despite the diverging task objectives: in our work, we detach our analysis from intents or factuality of question, providing a strict formulation of evasion strategies. To this end, unanswered false presuppositions are not necessarily connected to the intent to deceive. We made this selection not only in order to differentiate from \citet{ferracane-etal-2021-answer}, but also to restrict the large set of possible interpretations arising under varying intents. For example, a question containing a false premise, such as "Why is the earth flat?" accompanied with a response "The earth is not flat." does not receive the information requested -the reason \textit{why} the earth is flat- but rather utilizes a factual statement -the earth is \textit{scientifically proven} \textbf{not} to be flat- to form the response, which can be classified as an Ambivalent Reply. In case the question contained a valid statement (e.g. "Why is the earth round?") a similarly formatted reply ("The earth is not round") would be again classified as Ambivalent Reply in terms of the information provided, even though it reflects reduced factual knowledge or an intent to deceive from the interviewer's side. However, recognizing intents can be subjective and highly variable, while measuring the degree and the type of information provided, as in our work, formulates a more deterministic and strict framework. At the same time, we do not require detailed knowledge of the facts contained in the question, which may be unavailable even to audience with related background; a separate factuality analysis would reveal potential knowledge gaps highlighting possible interpretations of the question at hand. Overall, our annotated responses contain a specific label regardless the intent and the factuality of the question.

We will further analyze the performance of our models using the dataset referenced in \cite{ferracane-etal-2021-answer}. By applying our models to their dataset, we aim to assess their generalizability across varied contexts. It is important to note that while both datasets predominantly cover the political domain and include press conferences of U.S. Presidents, their formulations are markedly distinct. Specifically, the dataset in \cite{ferracane-etal-2021-answer} is defined by its goal to determine not only if respondents intend to answer questions but also if their responses are truthful. This subjective approach necessitates a multi-label problem framework where instances might receive conflicting labels, such as ``Can't answer Sincere'' and ``Can't answer Lying.'' This complexity arises when one annotator perceives deception, while another believes in the sincerity of the response. However, more complex situations may arise, such as when one annotator labels an instance as ``Answer'' and another labels it as ``Can't Answer - Lying.'' This variation indicates that differences in perceived intent and truthfulness can completely alter the label concerning the answerability of the response, contrary to expectations.

Contrastingly, our model's framework does not consider the intent or truthfulness of responses, focusing solely on whether the response addresses the question. Discrepancies in labeling by annotators are resolved by an expert, streamlining the process and ensuring each instance maintains a singular, clear label. This approach aligns with our primary objective: determining the direct answerability of responses, irrespective of underlying intentions or truthfulness.

\begin{figure*}[htb] 
  \centering 
  \begin{minipage}{0.49\textwidth}
    \includegraphics[width=\linewidth]{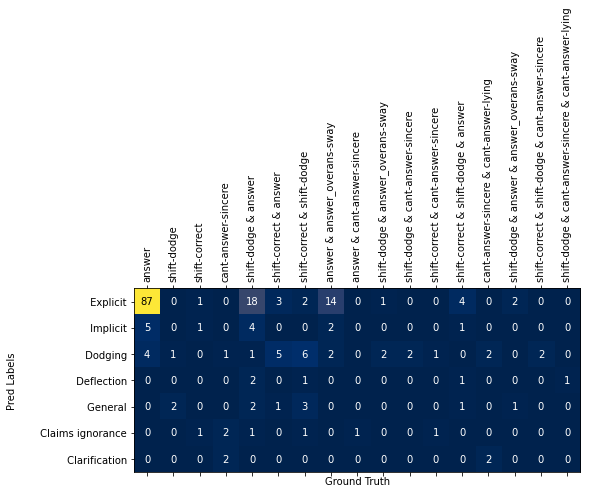} 
    \caption{Results of Llama-70b trained using the evasion based clarity for dev set of \cite{ferracane-etal-2021-answer}.}
    \label{fig:sub_dev_set}
  \end{minipage}\hfill 
  \begin{minipage}{0.49\textwidth}
    \includegraphics[width=\linewidth]{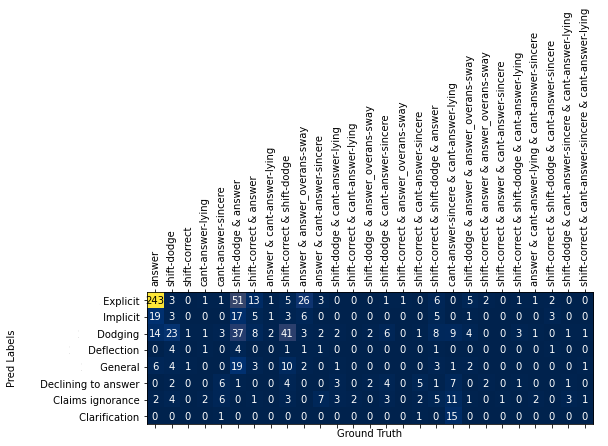} 
    \caption{Results of Llama-70b trained using the evasion based clarity for the training set of \cite{ferracane-etal-2021-answer}.}
    \label{fig:sub_train_set}
  \end{minipage}
\end{figure*}

Further, we seek to evaluate the efficacy of our top-performing model, trained on our dataset, on the dataset proposed in \cite{ferracane-etal-2021-answer}. Initially, we eliminate all duplicate entries, then process the remaining data through the Llama-70b model, which was trained using evasion-based direct clarity strategies. Figures \ref{fig:sub_dev_set} and \ref{fig:sub_train_set} illustrate the comparison between the ground truth and our predicted labels across the training and development sets. This comparison is crucial, especially considering the development set's relatively small size—it comprises fewer than 200 instances across 27 labels, with some labels lacking adequate representation.

Firstly, it is evident that this dataset is also highly unbalanced, with 'Answer' being the most frequently occurring label, similar to our own dataset. Additionally, there is a clear alignment between the predicted labels using our taxonomy and the ground truth labels. For instance, instances labeled with ``shift-dodge \& can't answer lying'' are predominantly classified under one of the corresponding labels from our taxonomy, such as ``Declining to answer,'' ``Claims ignorance,'' or ``Dodging.'' To provide a quantifiable measure of the model's performance across both tasks, we evaluate the model's effectiveness solely on instances that have a single ground truth label in both sets, as shown in Table \ref{tab:subj_results}, employing a weighted average strategy.

\begin{table}[]
\centering
\small
\begin{tabular}{c|cccc} 
\toprule
      & Acc. & Prec. & Recall & F1   \\ \midrule
Dev   & 0.85     & 0.89  & 0.85   & 0.87 \\
Train & 0.81     & 0.85  & 0.81   & 0.82 \\

\bottomrule
\end{tabular}
\caption{The performance of the Llama-70b trained using the evasion based clarity, on development and training sets.}
\label{tab:subj_results}
\end{table}

\begin{figure*}[htb] 
  \centering 
  \begin{minipage}{0.49\textwidth}
    \includegraphics[width=\linewidth]{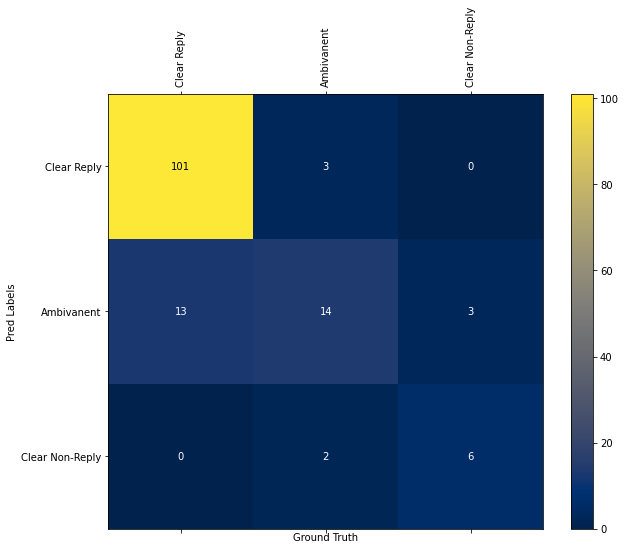} 
    \caption{Confusion matrix of Llama-70b trained using the evasion based clarity for dev set of \cite{ferracane-etal-2021-answer} for the single labelled instances.}
    \label{fig:conf_dev_set}
  \end{minipage}\hfill 
  \begin{minipage}{0.49\textwidth}
    \includegraphics[width=\linewidth]{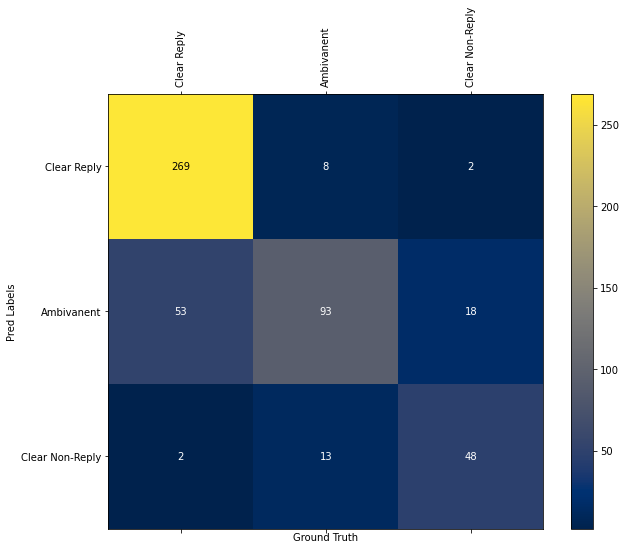} 
    \caption{Results of Llama-70b trained using the evasion based clarity for the training set of \cite{ferracane-etal-2021-answer} for the single labelled instances.}
    \label{fig:conf_train_set}
  \end{minipage}
\end{figure*}

The results indicate that our model can generalize effectively, performing well on a dataset annotated with a different strategy. However, it is important to note that the improved outcomes on this dataset, compared to our own, might be attributed to instances having clear and consistent answers across different annotators, suggesting a higher clarity in these instances. Finally, Figures \ref{fig:conf_dev_set} and \ref{fig:conf_train_set} display the confusion matrices comparing the ground truth with our results for instances with single labels.

\section{Prompting details}
\label{sec:prompting}

\paragraph{Prompt for generating sQAs}

The following prompt was provided to ChatGPT to obtain the sQAs of the multi-part pairs, as well as to request the appropriate label based on the proposed taxonomy.

\par\noindent\rule{0.5\textwidth}{0.4pt}
\small message\_0 = ``````

Point out what is this question Q asking. Stating of facts are not considered as questions, but only requests of information do. If it's a multi-part question, break down it the separate components that it asks. Use the following template to show the questions and the questions only.

The question consists of N parts: [add the correct N depending on the question]
[Enumerate the question parts and give each part a short title in the beginning of the line]
``````

``````

message\_1 = ``````

Now analyse the information that this answer provides, especially regarding the points being asked, filling the following template.

Template 
---
The response provides the following information regarding these points:
[Enumerate the question parts along with their title, followed by the relevant information given per part in the response]
---
Answer:

``````

message\_2 = ``````

For each part of the question, and the questions only,  use the following taxonomy to describe what type of a reply did the answer provide to it, along with a brief clarification for each choice. Note that if the question does not request elaboration, you should not consider the lack of elaboration in the answer as a lack of information. 
---
Template:

Question part: [number and title]

Verdict: [taxonomy code and title]

Explanation:

---

<taxonomy>

``````
\par\noindent\rule{0.5\textwidth}{0.4pt}
\normalsize

\paragraph{Prompt for generating counter-sQAs}

In addition to this prompt, we create some ``counter-sQAs`` to assess the annotators' reliance on the extracted sQAs rather than the original multi-part pairs as provided in the interviews. The following prompt was appended to the previous one:

\par\noindent\rule{0.5\textwidth}{0.4pt}
\small message\_3 = ``````

Now, try to create an QAs of the response to intentionally mislead someone into thinking that the answer corresponds to a different category than the one you initially predicted. For instance, if your prediction is 'Explicit,'' generate an sQA that could make someone believe it is a ``General'' response or any other label of your choice. The sQA should be at the same length as the original one. Start by selecting the counterlabel and then write the sQs using the following template:

Template 

---

The response provides the following information regarding these points:

[Enumerate the question parts along with:

- title

- original label

- counterfactual label 

- fake information for each part in the response supporting the counterfactual label.]

---

Answer:

``````
\par\noindent\rule{0.5\textwidth}{0.4pt}
\normalsize 

\paragraph{Zero-shot prompt for classification}

The following prompt was used for addressing the evasion problem in the zero-shot scenario.

\par\noindent\rule{0.5\textwidth}{0.4pt}
\small message\_0 = ``````
Based on a segment of the interview in which the interviewer poses a series of questions, classify the type of response provided by the interviewee for the following question using the following taxonomy and then provide a chain of thought explanation for your decision:
\\\\
<Taxonomy>
\\\\
You are required to respond with a single term corresponding to the Taxonomy code and only.
\\\\
\#\#\# Part of the interview \#\#\#
\\
<Part of the interview>
\\
\#\#\# Question \#\#\#
\\
<Question>
\\
Taxonomy code: 
``````
\par\noindent\rule{0.5\textwidth}{0.4pt}
\normalsize 

The following prompt was used for addressing the clarity problem in the zero-shot scenario.

\par\noindent\rule{0.5\textwidth}{0.4pt}
\small message\_0 = ``````
Based on a segment of the interview in which the interviewer poses a series of questions, classify the type of response provided by the interviewee for the following question using the following taxonomy and then provide a chain of thought explanation for your decision:
\\\\
1. Clear Reply - The information requested is explicitly stated (in the requested form) \\
2. Clear Non-Reply - The information requested is not given at all due to ignorance, need for clarification or declining to answer \\
3. Ambiguous - The information requested is given in an incomplete way e.g. the answer is too general, partial, implicit, dodging or deflection.
\\\\
You are required to respond with a single term corresponding to the Taxonomy code and only.
\\\\
\#\#\# Part of the interview \#\#\#
\\
<Part of the interview>
\\
\#\#\# Question \#\#\#
\\
<Question>
\\
Taxonomy code: 
``````

\par\noindent\rule{0.5\textwidth}{0.4pt}
\normalsize

\paragraph{Chain-of-Thought (CoT) prompt for classification}

The following prompt was used for addressing the evasion problem in the CoT scenario.

\par\noindent\rule{0.5\textwidth}{0.4pt}
\small message\_0 = ``````
Based on a segment of the interview in which the interviewer poses a series of questions, classify the type of response provided by the interviewee for the following question using the following taxonomy and then provide a chain of thought explanation for your decision:
\\\\
<Taxonomy>
\\\\
You are required to respond with a single term corresponding to the Taxonomy code as well as the chain of thought explanation.
\\\\
Let's think step by step. \\
\#\#\# Part of the interview \#\#\#
\\
<Part of the interview>
\\
\#\#\# Question \#\#\#
\\
<Question>
\\
Taxonomy code: 
``````
\par\noindent\rule{0.5\textwidth}{0.4pt}
\normalsize

The following prompt was used for addressing the clarity problem in the CoT scenario.

\par\noindent\rule{0.5\textwidth}{0.4pt}
\small message\_0 = ``````
Based on a segment of the interview in which the interviewer poses a series of questions, classify the type of response provided by the interviewee for the following question using the following taxonomy and then provide a chain of thought explanation for your decision:\\\\
1. Clear Reply - The information requested is explicitly stated (in the requested form)\\
2. Clear Non-Reply - The information requested is not given at all due to ignorance, need for clarification or declining to answer\\
3. Ambivalent - The information requested is given in an incomplete way e.g. the answer is too general, partial, implicit, dodging or deflection\\

You are required to respond with a single term corresponding to the Taxonomy code as well as the chain of thought explanation.
\\\\
Let's think step by step.\\
\#\#\# Part of the interview \#\#\#
\\
<Part of the interview>
\\
\#\#\# Question \#\#\#
\\
<Question>
\\
Taxonomy code: 
``````
\par\noindent\rule{0.5\textwidth}{0.4pt}
\normalsize 

\paragraph{Few-Shot (FS) prompt for classification}

The following prompt was used for addressing the evasion problem in the FS scenario.

\par\noindent\rule{0.5\textwidth}{0.4pt}
\small message\_0 = ``````
Based on a segment of the interview in which the interviewer poses a series of questions, classify the type of response provided by the interviewee for the following question using the following taxonomy:\\\\
<Taxonomy>
\\\\
Here is one small example for each term of the taxonony:

Question: \\
Do you have your own views about PR at Westminster don’t you?\\
Answer: \\
I do.\\
Label: Explicit \\
Explanation:  The answer directly gives the info requested. 

Question: Are you going to watch television? \\
Answer: What else is there to do?\\
Label: Implicit\\
Explanation: They suggest planning to watch TV, despite not explicitly stating it. 

Question: Do you like my new dress?\\
Answer: We are late.\\
Label: Dodging\\
Explanation: Does not even acknowledge the question and goes straight to another topic.

Question: Did you eat the last piece of pie?\\
Answer: I have to admit that this was a great recipe, I always like it when there are chocolate chips in the dough.\\
Label: Deflection\\
Explanation: Acknowledges the question but goes on a tangent about the chips, without answering. 

Question: Did you enjoy the film? \\
Answer: The directing was great.\\
Label: Partial/half-answer\\
Explanation: Directing is only part of what constitutes a film. 

Question: What's your favorite film? \\
Answer: Fight Club, Filth, and Hereditary.\\
Label: General\\
Explanation: The reply gives three movies instead of one, which makes the desired information unclear. 


Question: The hypothesis I was discussing, wouldn’t you regard that as a defeat?\\
Answer: I am not going to prophesy what will happen.\\
Label: Declining to answer\\
Explanation: Directly stating they won’t answer. 

Question: On what precise date did the government order the refit of the HMAS Kanimbla in preparation for its forward deployment to a possible war against Iraq? \\
Answer: I do not know that date. I will find out and let the House know. \\
Label: Claims ignorance \\
Explanation: Claims/admits they don’t have the information. 

Question: Was it your decision to release the fund? \\
Answer: You mean the public fund?\\
Label: Clarification\\
Explanation: Gives no data, asks for clarification. 


\#\#\# Part of the interview \#\#\#
\\
<Part of the interview>
\\
\#\#\# Question \#\#\#
\\
<Question>
\\
Taxonomy code: 
``````
\par\noindent\rule{0.5\textwidth}{0.4pt}
\normalsize

The following prompt was used for addressing the clarity problem in the FS scenario.

\par\noindent\rule{0.5\textwidth}{0.4pt}
\small message\_0 = ``````

Based on a segment of the interview in which the interviewer poses a series of questions, classify the type of response provided by the interviewee for the following question using the following taxonomy:

1. Clear Reply - The information requested is explicitly stated (in the requested form)\\
2. Clear Non-Reply - The information requested is not given at all due to ignorance, need for clarification or declining to answer\\
3. Ambivalent - The information requested is given in an incomplete way e.g. the answer is too general, partial, implicit, dodging or deflection\\

Here is one small example for each term of the taxonony:

Question: \\
Do you have your own views about PR at Westminster don’t you?\\
Answer: 
I do. \\
Label: Clear Reply \\
Explanation:  The answer directly gives the info requested.

Question: Are you going to watch television? \\
Answer: What else is there to do?\\
Label: Ambivalent\\
Explanation: They suggest planning to watch TV, despite not explicitly stating it.

Question: Do you like my new dress?\\
Answer: We are late. \\
Label: Ambivalent\\
Explanation: Does not even acknowledge the question and goes straight to another topic.

Question: Did you eat the last piece of pie?\\
Answer: I have to admit that this was a great recipe, I always like it when there are chocolate chips in the dough. \\
Label: Ambivalent \\
Explanation: Acknowledges the question but goes on a tangent about the chips, without answering.

Question: Did you enjoy the film?\\
Answer: The directing was great.\\
Label: Ambivalent\\
Explanation: Directing is only part of what constitutes a film.

Question: What's your favorite film? \\
Answer: Fight Club, Filth, and Hereditary. \\
Label: Ambivalent \\
Explanation: The reply gives three movies instead of one, which makes the desired information unclear.


Question: The hypothesis I was discussing, wouldn’t you regard that as a defeat? \\
Answer: I am not going to prophesy what will happen. \\
Label: Clear Non-Reply \\
Explanation: Directly stating they won’t answer.

Question: On what precise date did the government order the refit of the HMAS Kanimbla in preparation for its forward deployment to a possible war against Iraq? \\
Answer: I do not know that date. I will find out and let the House know. \\
Label: Clear Non-Reply \\
Explanation: Claims/admits they don’t have the information.

Question: Was it your decision to release the fund? \\
Answer: You mean the public fund? \\
Label: Clear Non-Reply \\
Explanation: Gives no data, asks for clarification.


\#\#\# Part of the interview \#\#\#
\\
<Part of the interview>
\\
\#\#\# Question \#\#\#
\\
<Question>
\\
Taxonomy code: 
``````
\par\noindent\rule{0.5\textwidth}{0.4pt}
\normalsize


\subsection{Prompt for LoRA fine-tuning}
\label{app:lora}

For the instruction-tuning part, we rely on LoRA fine-tuning \cite{hu2021lora} with  $r =16$, $alpha=32$ and $dropout=0.05$  using a subset of 2700 annotated samples as training set and the rest 750 as validation set. 
The following prompt was used for instruction-tuning, and it remained consistent across all models and the two methodologies (direct clarity and evasion-based clarity). The only distinction between the two different setups in the prompt was the specific label that the model should generate. Inference proceeded without sampling, though we did experiment with sampling, which resulted in slightly lower performance.

\par\noindent\rule{0.5\textwidth}{0.4pt}
\small message\_0 = ``````Based on a part of the interview where the interviewer asks a set of questions, classify the type of answer the interviewee provided for the following question\\
\#\#\# Part of the interview \#\#\#\\
<Interview Part>\\
\\\\
\#\#\# Question \#\#\#\\
<Question>\\
Label: <Label>\\
``````
\par\noindent\rule{0.5\textwidth}{0.4pt}
\normalsize

\section{Computational Resources}
All the experiments were conducted on a cluster with 4 NVIDIA A100-SXM4-40GB. The total hours of experimentation for training and inference (both for zero-shot and fine-tuned models) were 230 GPU hours and 440 CPU hours.

\end{document}

%% file: tables/rw_evasion_typology.tex
\begin{table}[h]
\small
\begin{enumerate}\setlength\itemsep{0em}
\hrule \vspace{0.1cm}
    \item \textbf{Ignores the question.} Makes no attempt to answer the question, or even to acknowledge it has been asked.
    \item \textbf{Acknowledges the question.} Acknowledges that a question has been asked, but equivocates.
    \item \textbf{Questions the question.} Requests clarification, or reflects the question back to the questioner.
    \item \textbf{Attacks the question.}
    \item \textbf{Personalisation.} Makes personal comments or attacks.
    \item \textbf{Declines to answer.}
    \item \textbf{Makes political points.}
    \item \textbf{Gives incomplete reply.}
    \item \textbf{Repeats answer to the previous question.}
    \item \textbf{States or implies has already answered the question.}
    \item \textbf{Apologises.}
    \item \textbf{Literalism.} The literal aspect of a question which was not intended to be taken literally is answered.
\end{enumerate}
\vspace{-0.2cm}
\hrule 
\vspace{0.1cm}
\caption{Equiv. typology by \citet{Bull2019CantAW}.}
\label{table:equivocation_typology}
\end{table}

%% file: tables/zeroshot.tex
\begin{table}[tb!]
\centering\renewcommand{\arraystretch}{1}
\small 
\begin{tabular}{P{1.1cm}P{2cm}P{0.5cm}P{0.5cm}P{0.5cm}P{0.5cm}}
\toprule
\small Classification variant & \small Model  & \small Acc. & \small Prec. & \small Recall & \small F1    \\ \midrule
\multicolumn{6}{c}{\small Prompting}\\
\midrule
\multirow{10}{4em}{\small direct clarity} &\small ZS Llama-70b  & \small  0.467  & \small  0.429   & \small 0.235	 &  \small 0.259 \\
&\small ZS Falcon-40b    & \small  0.240  & \small  0.252    & \small 0.247	 & \small 0.144 \\
&\small ZS ChatGPT        & \small  \underline{0.649}  & \small   \underline{0.476}   & \small  \underline{0.413}  & \small \underline{0.413}  \\\cmidrule{2-6}
&\small FS Llama-7b  & \small  0.23  & \small  0.159   & \small 0.474	 &  \small 0.219 \\ 
&\small FS Llama-13b  & \small  0.211  & \small  0.105   & \small 0.302	 &  \small 0.156 \\ 
&\small FS Llama-70b  & \small  \underline{0.667}  & \small  \underline{0.333}   & \small \underline{0.333}	 &  \small \underline{0.333} \\ 
&\small FS Falcon-7b    & \small  0.203  & \small  0.107    & \small 0.267	 & \small 0.152 \\
&\small FS Falcon-40b    & \small  0.29  & \small  0.13    & \small 0.336	 & \small 0.186 \\\cmidrule{2-6}
& \small standalone CoT      & \small 0.628  & \small  0.414	   & \small 0.376 	 & \small  0.368  \\ \midrule
\multirow{10}{4em}{\small evasion-based clarity} & \small ZS Llama-70b & \small 0.385  & \small  	0.396    & \small 0.308 & \small 0.261 \\
&\small ZS Falcon-40b  & \small 0.618   & \small  0.365    & \small 0.387 & \small 0.375 \\
& \small ZS ChatGPT           & \small \underline{0.640} & \small  \underline{0.507}  & \small  \underline{0.497} & \small  \underline{0.482}\\
\cmidrule{2-6} 
& \small FS Llama-7b & \small 0.274  & \small  	0.393    & \small 0.335 & \small  0.262 \\
& \small FS Llama-13b & \small 0.291  & \small  	0.452    & \small 0.363 & \small  0.259 \\
& \small FS Llama-70b & \small \underline{0.541}  & \small  	\underline{0.565}    & \small \underline{0.452} & \small  \underline{0.365} \\
&\small FS Falcon-7b  & \small 0.505   & \small  0.299    & \small 0.211 & \small 0.222 \\
&\small FS Falcon-40b  & \small 0.429   & \small  0.167    & \small 0.25 & \small 0.2 \\
\cmidrule{2-6}
&\small standalone CoT          & \small \underline{0.688} & \small  \underline{0.611}  & \small  \underline{0.514} & \small  \underline{0.510} \\ 
& \small multi CoT              & \small 0.549    & \small 0.459     & \small 0.500   & \small 0.462  \\
\midrule
\multicolumn{6}{c}{\small Tuned models}\\
\midrule
\multirow{8}{4em}{\small direct clarity} &  DeBERTa-base              & 0.58                     & 0.521                     & 0.453                      & 0.441                  \\
                                       & RoBERTa-base              & \underline{0.64}                    & \underline{0.579}                     & \underline{0.516}                      & \underline{0.53}                  \\
    
                                       & XLNet-base                & 0.694                    & 0.52                      & 0.523                      & 0.518                  \\ 
   \cmidrule{2-6}
&\small Llama-7b &  \small 0.489      & \small 0.452    & \small 0.529 & \small 0.457 \\
& \small Llama-13b     & \small 0.587    & \small 0.579      & \small 0.7  & \small 0.58 \\
& \small Llama-70b     & \small \underline{\textbf{0.759}}    & \small \underline{\textbf{0.67}}     & \small \underline{0.70}  & \small \underline{0.68} \\
&\small Falcon-7b   & \small 0.288   & \small 0.325    & \small 0.333 & \small 0.175 \\ 
&\small Falcon-40b   & \small 0.341   & \small 0.512    & \small 0.534 & \small 0.356 \\ \midrule

\multirow{8}{4em}{\small evasion-based clarity} & DeBERTa-base              & 0.555                    & 0.53                      & 0.671                      & 0.537                  \\

                                       & RoBERTa-base              & 0.577                    & 0.501                     & 0.534                      & 0.495                  \\
                                       & XLNet-base                & \underline{0.58}                   & \underline{0.523}                    & \underline{0.586}                   & \underline{0.546}\\
\cmidrule{2-6}
&\small Llama-7b    & \small 0.666    & \small 0.618    & \small 0.616 & \small 0.616 \\ 
&\small Llama-13b  & \small 0.675   & \small 0.617    & \small 0.616 & \small 0.616 \\ 
&\small Llama-70b  & \small \underline{0.713}   & \small \underline{\textbf{0.67}}    & \small \underline{\textbf{0.71}} & \small \underline{\textbf{0.682}} \\ 
&\small Falcon-7b   & \small 0.533   & \small 0.429    & \small 0.386 & \small 0.397 \\ 
&\small Falcon-40b   & \small 0.621   & \small 0.616    & \small 0.532 & \small 0.558 \\ 
\bottomrule
\end{tabular}
\caption{Results for ZS, FS \& CoT prompting inference, as well as for fine/instruction-tuned models. The best results for each prompting/training variant are \underline{underlined} and best results overall are also in \textbf{bold}.}
\label{tab:llm-results-zs}
\end{table}
\vspace{-0cm}

%% file: tables/fewshot.tex
\begin{table}[tb!]
\centering
\begin{tabular}{P{1.2cm}P{1.5cm}P{0.5cm}P{0.5cm}P{0.5cm}P{0.5cm}}
\toprule
\small Classification variant & \small Model  & \small Acc. & \small Prec. & \small Recall & \small F1    \\ \midrule
\multirow{5}{4em}{\small direct clarity} 
&\small Llama-7b  & \small  0.23  & \small  0.159   & \small 0.474	 &  \small 0.219 \\ 
&\small Llama-13b  & \small  0.211  & \small  0.105   & \small 0.302	 &  \small 0.156 \\ 
&\small Llama-70b  & \small  \textbf{0.667}  & \small  \underline{0.333}   & \small \underline{0.333}	 &  \small \underline{0.333} \\ 
&\small Falcon-7b    & \small  0.203  & \small  0.107    & \small 0.267	 & \small 0.152 \\
&\small Falcon-40b    & \small  0.29  & \small  0.13    & \small 0.336	 & \small 0.186 \\ 
\midrule
\multirow{5}{4em}{\small evasion-based clarity} & 
\small Llama-7b & \small 0.274  & \small  	0.393    & \small 0.335 & \small  0.262 \\
& \small Llama-13b & \small 0.291  & \small  	0.452    & \small 0.363 & \small  0.259 \\
& \small Llama-70b & \small \underline{0.541}  & \small  	\textbf{0.565}    & \small \textbf{0.452} & \small  \textbf{0.365} \\
&\small Falcon-7b  & \small 0.505   & \small  0.299    & \small 0.211 & \small 0.222 \\
&\small Falcon-40b  & \small 0.429   & \small  0.167    & \small 0.25 & \small 0.2 \\
 \bottomrule
\end{tabular}
\caption{Classification results for few-shot (FS) inference. The best results for each strategy are \underline{underlined} and best results overall are also in \textbf{bold}.}
\label{tab:llm-results-fs}
\end{table}


%% file: tables/multi_questions_lora.tex
\begin{table}[]
\small
\centering
\begin{tabular}{P{1.2cm}P{1.5cm}P{0.5cm}P{0.6cm}P{0.6cm}P{0.6cm}} \toprule

                               Classification variant          & Model                        & Acc. & Prec. & Recall & F1             \\   \midrule

\multirow{10}{4em}{direct clarity} & \multirow{2}{*}{Llama-7b}  & 0.47     & 0.403     & 0.48   & 0.402          \\
                                        &                              & 0.53     & 0.537     & 0.588  & \textbf{0.538} \\
                                        & \multirow{2}{*}{Llama-13b} & 0.59     & 0.547     & 0.711  & 0.548          \\
                                        &                              & 0.59     & 0.625     & 0.694  & \textbf{0.621} \\
                                        & \multirow{2}{*}{Llama-70b} & 0.74     & 0.594     & 0.648  & 0.612          \\
                                        &                              & 0.78     & 0.705     & 0.742  & \textbf{0.72}  \\
                                        & \multirow{2}{*}{Falcon-7b}  & 0.25     & 0.319     & 0.337  & 0.158          \\
                                        &                              & 0.37     & 0.341     & 0.329  & \textbf{0.21}  \\
                                        & \multirow{2}{*}{Falcon-40b} & 0.29     & 0.432     & 0.468  & 0.284          \\
                                        &                              & 0.44     & 0.67      & 0.629  & \textbf{0.459}
\\ \midrule
\multirow{10}{4em}{evasion-based clarity}   
& \multirow{2}{*}{Llama-7b}  & 0.67     & 0.593     & 0.59   & 0.591          \\
                                        &                              & 0.64     & 0.602     & 0.622  & \textbf{0.608}          \\
                                        & \multirow{2}{*}{Llama-13b} & 0.69     & 0.592     & 0.581  & 0.586          \\
                                        &                              & 0.64     & 0.635     & 0.679  & \textbf{0.654} \\
                                        & \multirow{2}{*}{Llama-70b} & 0.7      & 0.601     & 0.656  & 0.62           \\
                                        &                              & 0.73     & 0.75      & 0.785  & \textbf{0.761} \\
                                        & \multirow{2}{*}{Falcon-7b}  & 0.54     & 0.442     & 0.372  & 0.384          \\
                                        &                              & 0.52     & 0.429     & 0.413  & \textbf{0.418} \\
                                        & \multirow{2}{*}{Falcon-40b} & 0.64     & 0.62      & 0.47   & 0.493          \\
                                        &                              & 0.58     & 0.578     & 0.598  & \textbf{0.586} 
\\ \bottomrule
\end{tabular}
\caption{Classification results for instruction-tuned models. The best results overall are in \textbf{bold}. The first line of each model shows the results for the set containing only multi-part questions, while the second line shows the results for single-part questions.}
\label{tab:loras_single}
\end{table}

%% file: tables/multi_questions_gpt.tex
\begin{table}[h!]
\small
\centering
\begin{tabular}{P{1.2cm}P{1.5cm}P{0.5cm}P{0.6cm}P{0.6cm}P{0.6cm}} 
\toprule

                             Classification variant            & Model                        & Acc. & Prec. & Recall & F1             \\   \midrule

\multirow{4}{4em}{

\small direct clarity} & \multirow{2}{4em}{zero-shot}      & 0.668 & 0.418 & 0.37 & 0.37 \\
                            &    & \textbf{0.625} & \textbf{0.559} & \textbf{0.483}  & \textbf{0.461}  \\\cmidrule{2-6}
& \multirow{2}{4em}{standalone CoT} & 0.649 & 0.347 & 0.34 & 0.332 \\
                              &  & \textbf{0.607} & \textbf{0.537} & \textbf{0.441}  & \textbf{0.418} \\ \midrule
\multirow{4}{4em}{
\small evasion based clarity}   
&\multirow{2}{4em}{zero-shot}      & 0.639 & 0.443 & 0.442 & 0.436  \\
                              &  & \textbf{0.661} & \textbf{0.683} & \textbf{0.603} & \textbf{0.56} \\\cmidrule{2-6}
& \multirow{2}{4em}{standalone CoT} & 0.712 & 0.568 & 0.483 & 0.489 \\
                             &   & \textbf{0.643} & \textbf{0.657} & \textbf{0.558} & \textbf{0.536} \\
 \bottomrule
\end{tabular}
\caption{Classification results for ChatGPT using zero-shot and chain-of-thought inference for  the two subsets (single- and multi-part questions). The best results for each subset are in \textbf{bold}. The first line of each model shows the results for the set containing only multi-part questions, while the second line shows the results for single-part questions.}
\label{tab:gpt_single}
\end{table}

%% file: tables/knowledgeability.tex
\begin{table}[]
\small
\centering
\begin{tabular}{P{1.2cm}P{1.5cm}P{0.5cm}P{0.6cm}P{0.6cm}P{0.6cm}}
\toprule
               Classification variant    & Model                       & Acc.  & Prec. & Recall & F1    \\ \toprule
\multirow{10}{4em}{direct clarity} & \multirow{2}{*}{Llama-7b}   & 0.434 & 0.375 & 0.439  & 0.369 \\
                   &                             & \textbf{0.57}  & \textbf{0.56}  & \textbf{0.62}1  & \textbf{0.565} \\
                   & \multirow{2}{*}{Llama-13b}  & 0.55  & 0.527 & 0.663  & 0.52  \\
                   &                             & \textbf{0.639} & \textbf{0.631} & \textbf{0.731}  & \textbf{0.638} \\
                   & \multirow{2}{*}{Llama-70b}  & 0.752 & 0.65  & 0.777  & 0.69  \\
                   &                             & \textbf{0.768} & \textbf{0.7  } & \textbf{0.686}  & \textbf{0.692} \\
                   & \multirow{2}{*}{Falcon-7b}  & 0.266 & 0.255 & 0.319  & 0.148 \\
                   &                             & \textbf{0.32 } & \textbf{0.348} & \textbf{0.355}  & \textbf{0.213} \\
                   & \multirow{2}{*}{Falcon-40b} & 0.328 & 0.489 & 0.504  & 0.331 \\
                   &                             & \textbf{0.359} & \textbf{0.533} & \textbf{0.55 }  & \textbf{0.374} \\ \midrule
\multirow{10}{4em}{evasion based clarity} & \multirow{2}{*}{Llama-7b}   & 0.635 & 0.57  & 0.557  & 0.563 \\
                   &                             & \textbf{0.711} & \textbf{0.67} & \textbf{0.678}  & \textbf{0.673} \\
                   & \multirow{2}{*}{Llama-13b}  & 0.651 & 0.573 & 0.611  & 0.589 \\
                   &                             & \textbf{0.711} & \textbf{0.674} & \textbf{0.636}  & \textbf{0.653} \\
                   & \multirow{2}{*}{Llama-70b}  & 0.709 & 0.637 & 0.706  & 0.661 \\
                   &                             & \textbf{0.719} & \textbf{0.701} & \textbf{0.718}  & \textbf{0.702} \\
                   & \multirow{2}{*}{Falcon-7b}  & 0.497 & 0.387 & 0.319  & 0.332 \\
                   &                             & \textbf{0.586} & \textbf{0.488} & \textbf{0.473}  & \textbf{0.473} \\
                   & \multirow{2}{*}{Falcon-40b} & 0.598 & 0.531 & 0.45   & 0.468 \\
                   &                             & \textbf{0.656} & \textbf{0.665} & \textbf{0.601}  & \textbf{0.622} \\ \bottomrule
\end{tabular}
\caption{Classification results for instruction-tuned models. The best results overall are in \textbf{bold}. The first line of each model shows the results for the subset consisting exclusively of instances that contain named entities, while the second line shows the results for the subset without named entities.}
\label{tab:ner_lora}
\end{table}

%% file: tables/knowledgeability_gpt.tex
\begin{table}[]
\small
\centering
\begin{tabular}{P{1.2cm}P{1.5cm}P{0.5cm}P{0.6cm}P{0.6cm}P{0.6cm}} 
\toprule

                                  Classification variant      & Model                        & Acc. & Prec. & Recall & F1             \\   \midrule

\multirow{4}{4em}{

\small direct clarity} & \multirow{2}{4em}{zero-shot}      & 0.651 & 0.416 & 0.371 & 0.354 \\
                            &    & \textbf{0.641} & \textbf{0.53 }& \textbf{0.449}  & \textbf{0.463}  \\\cmidrule{2-6}
& \multirow{2}{4em}{standalone CoT} & 0.614 & 0.333 & 0.326 & 0.311 \\
                              &  & \textbf{0.648} & \textbf{0.518} & \textbf{0.429}  & \textbf{0.434} \\ \midrule
\multirow{4}{4em}{evasion based clarity}   
&\multirow{2}{4em}{zero-shot}      & 0.635 & 0.457 & 0.44 & 0.42  \\
                              &  & \textbf{0.648} & \textbf{0.559} & \textbf{0.532} & \textbf{0.536} \\\cmidrule{2-6}
& \multirow{2}{4em}{standalone CoT} & 0.712 & 0.568 & 0.483 & 0.489 \\
                             &   & \textbf{0.677} & \textbf{0.657} & \textbf{0.535} & \textbf{0.551} \\
 \bottomrule
\end{tabular}
\caption{Knowledge-related classification results for ChatGPT using zero-shot and chain-of-thought inference for  the two subset. The best results for each subset are in \textbf{bold}. The first line of each model shows the results for the subset consisting exclusively of instances that contain named entities, while the second line shows the results for the subset without named entities.}
\label{tab:ner_gpt}
\end{table}

%% file: tables/evasion_classification.tex
\begin{table}[t!]
\centering
\small
\begin{tabular}{lcccc}
\toprule
Model      & Acc.  & Prec. & Recall & F1    \\ \toprule
LLama-7b   & 0.454 & 0.498  & 0.458  & 0.444 \\
LLama-13b  & 0.464 & 0.429  & 0.49   & 0.423 \\
LLama-70b  & \textbf{0.571} & \textbf{0.571}  & \textbf{0.558}  & \textbf{0.545} \\
Falcon-7b  & 0.363 & 0.226  & 0.216  & 0.212 \\
Falcon-40b & 0.476 & 0.558  & 0.475  & 0.492 \\ \bottomrule
\end{tabular}
\caption{ Classification results for instruction-tuned models for the evasion classification. The best results are in \textbf{bold}.}
\label{tab:lora_evasion_classification}
\end{table}

\begin{table}[t!]
\centering
\small
\begin{tabular}{lcccc}
\toprule
Model      & Acc.  & Prec. & Recall & F1    \\ \toprule
zero-shot   & \textbf{0.315} & 0.266  & \textbf{0.284}  & \textbf{0.244} \\
standalone CoT  & 0.259 & \textbf{0.293}  & 0.279   & 0.229 \\
 \bottomrule
\end{tabular}
\caption{ Classification results for evasion classification using zero-shot and chain-of-thought for prompting chatGPT which is best performing model using only prompting techniques. The best results are in \textbf{bold}.}
\label{tab:gpt_evasion_classification}

\end{table}

%% file: tables/encoders_ins.tex
\begin{table}[]
\small
\centering
\begin{tabular}{P{1cm}P{1.8cm}P{0.5cm}P{0.6cm}P{0.6cm}P{0.6cm}}
\toprule
         Classification variant            & Model                         & Acc.  & Prec. & Recall  & F1    \\ \midrule
\multirow{6}{4em}{direct clarity}        & \multirow{2}{*}{DebERTa-base} & \textbf{0.562} & \textbf{0.521}     & \textbf{0.467} & \textbf{0.465} \\
                                       &                               & 0.593 & 0.512     & 0.439 & 0.416 \\
                                       & \multirow{2}{*}{RoBERTa-base} & \textbf{0.625} & \textbf{0.614}     & \textbf{0.593} & \textbf{0.592} \\
                                       &                               & 0.651 & 0.383     & 0.405 & 0.392 \\
                                       & \multirow{2}{*}{XLNet-base}   & \textbf{0.68 } & \textbf{0.557}     & \textbf{0.571} & \textbf{0.56 } \\
                                       &                               & 0.704 & 0.481     & 0.468 & 0.472 \\ \midrule
\multirow{6}{4em}{evasion based clarity} & \multirow{2}{*}{DebERTa-base} & \textbf{0.57 } & \textbf{0.576}     & \textbf{0.645} & \textbf{0.568} \\
                                       &                               & 0.545 & 0.498     & 0.715 & 0.509 \\
                                       & \multirow{2}{*}{RoBERTa-base} & \textbf{0.539} & \textbf{0.55 }     & \textbf{0.581} & \textbf{0.543} \\
                                       &                               & 0.603 & 0.401     & 0.439 & 0.397 \\
                                       & \multirow{2}{*}{XLNet-base}   & \textbf{0.594} & \textbf{0.552}     & \textbf{0.617} & \textbf{0.574} \\
                                       &                               & 0.571 & 0.49      & 0.541 & 0.51  \\
 \bottomrule
\end{tabular}
\caption{Classification results for encoders. The best results overall are in \textbf{bold}. The first line of each model shows the results for the set containing only interview parts that contains named entities, while the second line shows the results for the parts withouts named entities.}
\label{tab:ner_enc}
\end{table}

\begin{table}[]
\small
\centering
\begin{tabular}{P{1cm}P{1.8cm}P{0.5cm}P{0.6cm}P{0.6cm}P{0.6cm}}
\toprule
 Classification variant            & Model                         & Acc.  & Prec. & Recall  & F1       \\ \midrule
\multirow{6}{4em}{direct clarity}        & \multirow{2}{*}{DebERTa-base} & \textbf{0.615} & \textbf{0.508}     & \textbf{0.469} & \textbf{0.44 } \\
                                       &                               & 0.518 & 0.538     & 0.438 & 0.429 \\
                                       & \multirow{2}{*}{RoBERTa-base} & 0.629 & 0.482     & 0.437 & 0.438 \\
                                       &                               & \textbf{0.661} & \textbf{0.649}     & \textbf{0.595} & \textbf{0.612} \\
                                       & \multirow{2}{*}{XLNet-base}   & 0.702 & 0.45      & 0.453 & 0.442 \\
                                       &                               & \textbf{0.679} & \textbf{0.626}     & \textbf{0.588} & \textbf{0.604} \\ \midrule
\multirow{6}{4em}{evasion based clarity} & \multirow{2}{*}{DebERTa-base} & 0.576 & 0.492     & 0.685 & 0.51  \\
                                       &                               & \textbf{0.518} & \textbf{0.624}     & \textbf{0.64 } & \textbf{0.563} \\
                                       & \multirow{2}{*}{RoBERTa-base} & 0.561 & 0.369     & 0.4   & 0.361 \\
                                       &                               & \textbf{0.607} & \textbf{0.618}     & \textbf{0.651} & \textbf{0.613} \\
                                       & \multirow{2}{*}{XLNet-base}   & 0.527 & 0.413     & 0.479 & 0.43  \\
                                       &                               & \textbf{0.679} & \textbf{0.707}     & \textbf{0.706} & \textbf{0.706} \\
 \bottomrule
\end{tabular}
\caption{Classification results for encoders. The best results overall are in \textbf{bold}. The first line of each model shows the results for the set containing only multi-part questions, while the second line shows the results for single-part questions.}
\label{tab:single_enc}
\end{table}

%% file: tables/encoders_results.tex
\begin{table}[tb!]
\centering
\small
\begin{tabular}{P{1cm}P{2.2cm}P{0.5cm}P{0.5cm}P{0.5cm}P{0.5cm}}
\toprule
\small Classification variant & \small Model  & \small Acc. & \small Prec. & \small Recall & \small F1    \\ \midrule

\multirow{5}{4em}{direct clarity}        & DeBERTa-base              & 0.58                     & 0.521                     & 0.453                      & 0.441                  \\
                                       & DeBERTa-large             & 0.691                    & 0.23                      & 0.333                      & 0.272                  \\
                                       & RoBERTa-base              & \underline{0.64}                    & \underline{0.579}                     & \underline{0.516}                      & \underline{0.53}                  \\
                                       & RoBERTa-large             & 0.593                    & 0.198                     & 0.333                      & 0.248                  \\
                                       & XLNet-base                & 0.694                    & 0.52                      & 0.523                      & 0.518                  \\ 
                                       & XLNet-large & 0.565 &	0.188	& 0.333	& 0.241 \\
                                       
                                        \midrule
\multirow{5}{4em}{evasion based clarity} & DeBERTa-base              & 0.555                    & 0.53                      & 0.671                      & 0.537                  \\
                                       & DeBERTa-large             & 0.249                    & 0.083                     & 0.333                      & 0.133                  \\
                                       & RoBERTa-base              & 0.577                    & 0.501                     & 0.534                      & 0.495                  \\
                                       & RoBERTa-large             & 0.278                    & 0.093                     & 0.333                      & 0.145                  \\
                                       & XLNet-base                & \underline{\textbf{0.58}}                    & \underline{\textbf{0.523}}                     & \underline{\textbf{0.586}}                    & \underline{\textbf{0.546}} \\ 
                                       
                                       & XLNet-large & 0.385	& 0.128	& 0.333	& 0.185 \\
                                       
                                       \bottomrule                 
\end{tabular}
\caption{Classification results for fine-tuned encoder models on the test set. The best results for each strategy are \underline{underlined} and best results overall are also in \textbf{bold}.}
\label{tab:encoder-results}
\end{table}

\begin{table}[tb!]
\centering
\small
\begin{tabular}{P{1cm}P{2.2cm}P{0.5cm}P{0.5cm}P{0.5cm}P{0.5cm}}
\toprule
\small Classification variant & \small Model  & \small Acc. & \small Prec. & \small Recall & \small F1    \\ \midrule

\multirow{7}{4em}{direct clarity}        & DeBERTa-base &	0.572 &	0.548 &	0.469 &	0.469 \\
& DeBERTa-large &	0.647 &	0.216	& 0.333 &	0.262 \\
& RoBERTa-base &	0.595 &	0.569 &	0.524 &	0.524 \\
& RoBERTa-large &	0.566 &	0.189 &	0.333 &	0.241 \\
& Llama-7b &	0.506	& 0.49	& 0.529	& 0.495 \\
& Llama-13b &	0.673	& 0.657	& 0.74	& 0.67 \\
& Llama-70b &	\textbf{\underline{0.775}}	& \textbf{\underline{0.743}}	 & \textbf{\underline{0.724}} &	\textbf{\underline{0.732}} \\

                                        \midrule
\multirow{7}{4em}{evasion based clarity} & 
DeBERTa-base &	0.561 &	0.568 &	0.664 &	0.569 \\
& DeBERTa-large &	0.254 &	0.085 &	0.333 & 	0.135 \\
& RoBERTa-base &	0.555	& 0.538 & 	0.548 &	0.512 \\
& RoBERTa-large & 0.277 &	0.092	& 0.333 &	0.145 \\
& Llama-7b	& 0.678	& 0.651	& 0.624	& 0.633 \\
& Llama-13b & 0.707	& 0.692 &	0.646 &	0.665 \\
& Llama-70b  & \underline{0.724}	 & \underline{0.695}	 & \underline{0.702}	 & 
\underline{0.698}
                                       \\ \bottomrule                 
\end{tabular}
\caption{Classification results for fine-tuned encoder models on the 173 samples of the test set that the input was not truncated. For comparison reasons the table is also depicted the performance of the instruction tuned LLama for this subset. The best results for each strategy are \underline{underlined} and best results overall are also in \textbf{bold}.}
\label{tab:encoder-not-truncated-results}
\end{table}

\begin{table}[tb!]
\centering
\small
\begin{tabular}{P{1cm}P{2.2cm}P{0.5cm}P{0.5cm}P{0.5cm}P{0.5cm}}
\toprule
\small Classification variant & \small Model  & \small Acc. & \small Prec. & \small Recall & \small F1    \\ \midrule

\multirow{5}{4em}{direct clarity}        & DeBERTa-base &	0.59  & 0.381  &  0.343  &  0.309 \\
& DeBERTa-large &	0.743  & 0.248  &  0.333  &  0.284 \\
& RoBERTa-base &	\underline{0.694}  & \underline{0.403}  &  \underline{0.41 } &  \underline{0.406} \\
& RoBERTa-large &	0.625  & 0.208  &  0.333  &  0.256 \\

                                        \midrule
\multirow{5}{4em}{evasion based clarity} & 
DeBERTa-base &	\textbf{\underline{0.549}}  & \textbf{\underline{0.44 }} &  \textbf{\underline{0.734}}  &  \textbf{\underline{0.424}} \\
& DeBERTa-large &	0.243  & 0.081  &  0.333  &  0.13 \\
& RoBERTa-base &	0.604  & 0.392  &  0.404  &  0.383 \\
& RoBERTa-large & 0.278  & 0.093  &  0.333  &  0.145
                                       \\ \bottomrule                 
\end{tabular}
\caption{Classification results for fine-tuned encoder models on the 144 samples of the test set that the input was truncated. The best results for each strategy are \underline{underlined} and best results overall are also in \textbf{bold}.}
\label{tab:encoder-truncated-results}

\end{table}